\documentclass{article}

\usepackage{microtype}
\usepackage{graphicx}
\usepackage{subfigure}
\usepackage{booktabs} 

\usepackage{hyperref}

\usepackage{makecell}

\usepackage{amsmath}
\usepackage{amssymb}
\usepackage{multirow}
\usepackage{hhline}
\usepackage[titlenumbered,ruled]{algorithm2e}
\usepackage{algorithmic}

\usepackage{footnote}
\makesavenoteenv{algorithm}

\usepackage{wrapfig}

\renewcommand{\algorithmiccomment}[1]{\hfill{$\triangleright$ #1}}

\usepackage[accepted]{icml2020}

\icmltitlerunning{AutoGAN-Distiller: Searching to Compress Generative Adversarial Networks}

\begin{document}

\twocolumn[
\icmltitle{AutoGAN-Distiller: Searching to Compress Generative Adversarial Networks}



\icmlsetsymbol{equal}{*}

\begin{icmlauthorlist}
\icmlauthor{Yonggan Fu}{rice}
\icmlauthor{Wuyang Chen}{tamu}
\icmlauthor{Haotao Wang}{tamu}
\icmlauthor{Haoran Li}{rice}
\icmlauthor{Yingyan Lin}{rice}
\icmlauthor{Zhangyang Wang}{tamu}
\end{icmlauthorlist}

\icmlaffiliation{rice}{Rice University, Houston, Texas, USA}
\icmlaffiliation{tamu}{Texas A\&M University, College Station, Texas, USA}

\icmlcorrespondingauthor{Zhangyang Wang}{atlaswang@tamu.edu }
\icmlcorrespondingauthor{Yingyan Lin}{yingyan.lin@rice.edu}

\icmlkeywords{Machine Learning, ICML}

\vskip 0.3in
]



\printAffiliationsAndNotice{}  

\begin{abstract}

The compression of Generative Adversarial Networks (GANs) has lately drawn attention, due to the increasing demand for deploying GANs into mobile devices for numerous applications such as image translation, enhancement and editing. However, compared to the substantial efforts to compressing other deep models, the research on compressing GANs (usually the generators) remains at its infancy stage. Existing GAN compression algorithms are limited to handling specific GAN architectures and losses. Inspired by the recent success of AutoML in deep compression, we
introduce AutoML to GAN compression and develop an \textit{AutoGAN-Distiller} (\textbf{AGD}) framework. Starting with a specifically designed efficient search space, AGD performs an end-to-end discovery for new efficient generators, given the target computational resource constraints. The search is guided by the original GAN model via knowledge distillation, therefore fulfilling the compression. AGD is \textbf{fully automatic}, \textbf{standalone} (i.e., needing no trained discriminators), and \textbf{generically applicable} to various GAN models. We evaluate AGD in two representative GAN tasks: image translation and super resolution. Without bells and whistles, AGD yields remarkably lightweight yet more competitive compressed models, that largely outperform existing alternatives. Our codes and pretrained models are available at: \url{https://github.com/TAMU-VITA/AGD}.
\end{abstract}
\section{Introduction}

Generative adversarial networks (GANs)~\cite{goodfellow2014generative,zhu2017unpaired} nowadays have empowered many real applications such as image stylization, image editing and enhancement. Those applications have driven the growing demand to deploy GANs, usually their trained \textbf{generators}, on resource-constrained platforms, e.g., for real-time style transfer and super resolution~\cite{shi2016real} in mobile APPs. However, just like other deep models, GAN generators require heavy memory and computation resources to run, which challenge most mobile devices. For example, the renowned image-to-image translation network, CycleGAN~\cite{zhu2017unpaired}, would cost 54 GFLOPs to process one 256$\times$256 image. For most mobile phones, that latency would notably degrade user experience, if ever feasible.

To reduce this gap, it is a natural motive to refer to model compression techniques \cite{han2015deep}. The current mainstream of model compression focuses on deep image classification or segmentation, which, as demonstrated in \cite{shu2019co}, cannot be easily extended to compress GANs (by default, defined as compressing their trained generators), That is mainly due to the vulnerability of their learned mappings (i.e., between high-dimensional structured image spaces), and the notoriously training instability (that challenges the re-training practice in compression). To our best knowledge, \textit{at the time of submission}, \cite{shu2019co} is the only published algorithm in GAN compression. They specifically focus on GANs with the cycle-consistency loss, and simultaneously compress generators of both directions. Despite the promising performance, their algorithm cannot be straightforwardly extended to other popular GAN types, such as encoder-decoder GANs \cite{wang2018esrgan,chen2018cartoongan} which are widely used in image editing/enhancement. Moreover, the proposed co-evolution algorithm was built on pruning only, while many other compression techniques, such as quantization \cite{jacob2018quantization}, knowledge distillation \cite{polino2018model}, and even AutoML \cite{he2018amc}, were left unexplored. Besides, their algorithm’s loss function relied on the original trained discriminator to be available. Yet in practice, this might not always be realistic: as most applications only use trained generators, the discriminators might often have been
discarded or no longer available
after training.

This paper aims to \textit{significantly push forward the application frontier of GAN compression}. For this new problem, we extensively refer to the state-of-the-art compression techniques including AutoML and knowledge distillation, while striving for more general techniques that can compress any common GAN generator. The resulting framework, dubbed \textit{AutoGAN-Distiller} (\textbf{AGD}), is the first AutoML framework
dedicated to GAN compression (alongside the concurrent work \cite{li2020gan}), and is also among a few earliest works that explore AutoML for GANs \cite{gong2019autogan}. AGD is established on a specifically designed search space of efficient generator building blocks, leveraging knowledge from state-of-the-art GANs for different tasks. It then performs differentiable neural architecture search under the target compression ratio (computational resource constraint), which preserves the original GAN generation quality via the guidance of knowledge distillation \cite{polino2018model}. AGD makes no assumption towards GAN structures, loss forms, nor even the availability of trained discriminators. It is therefore broadly applicable to compressing both CycleGANs and other GANs without the cycle loss. We demonstrate AGD on two representative mobile-based GAN applications: unpaired image translation (using a CycleGAN), and super resolution (using an encoder-decoder GAN). In both tasks, AGD overwhelms the state-of-the-arts \cite{li2016pruning,shu2019co,ledig2017photo,kim2016accurate}.

\section{Related Work}
\subsection{AutoML: Neural Architecture Search}
As one of the most significant sub-fields of AutoML \cite{hutter2019automated}, Neural Architecture Search (NAS)~\cite{zoph2016neural} seeks an optimal neural network architecture from data, instead of using hand-crafting. Already by now, NAS methods have outperformed manually designed architectures on a range of tasks such as image classification~\cite{tan2019efficientnet, tan2019mnasnet, howard2019searching, liu2018darts}, and segmentation~\cite{chen2018searching, liu2019auto, chen2019fasterseg}. A comprehensive survey of NAS could be found in \cite{elsken2018neural}

Introducing NAS to GANs appears to more challenging due to the training instability and lack of direct performance measures. \cite{gong2019autogan} developed the first NAS framework for GANs, on the task of unconditional image generation from random noise. It adopted a multi-level search strategy with an RNN controller to search for the generator architecture using reinforcement learning. However, the existing NAS for GAN framework is \textbf{not directly applicable} in our case due to the following reasons:
\vspace{-0.5em}
\begin{itemize}
    \item The framework in \cite{gong2019autogan} searched for a GAN from scratch, without referring to a trained model. There is also no computational resource constraint in their search. It is hence not a ``compression'' technique and lacks the ability to distill knowledge from those already-trained, high-performance GANs. 
    \item \cite{gong2019autogan} only demonstrated to find relatively small GAN models that synthesize low-resolution images from noise (e.g., a 6-layer generator on 32 $\times$ 32 CIFAR-10 images), and also did not consider image-to-image GANs. In comparison, the desired GAN compression shall apparently be able to handle the heavily-parameterized GANs on high-resolution images, with image-to-image applications (stylization, editing, enhancement, etc) as the primary focuses.
\end{itemize}

\vspace{-1em}
\subsection{Model Compression}
With the unprecedented demand for deploying deep models on resource-constrained platforms, many compression methods are proposed to narrow the gap between the model complexity and on-device resources \cite{han2020ghostnet,energynet, chen2020frequency}. Popular techniques include knowledge distillation~\cite{polino2018model}, pruning~\cite{han2015deep} and quantization~\cite{jacob2018quantization}. 

Knowledge distillation was proposed in \cite{hinton2015distilling} to transfer the knowledge from one model to another by imitating the soft labels generated by the former. It was widely applied in learning a compact network by fitting the soft labels generated by a pretrained large model~\cite{lopez2015unifying,bulo2016dropout,wang2018adversarial,chen2019data, chen2020distilling}, therefore compressing the latter.

Pruning \cite{han2015deep} refers to sparsifying the model weights by thresholding the unimportant weights. All of them follow a similar workflow: (iteratively) pruning the model to a smaller size and then retraining to recover accuracy. Structured pruning, e.g. channel pruning~\cite{wen2016learning,he2017channel}, was widely adopted as a hardware-friendly compression means \cite{gui2016feature}. 
\cite{luo2017thinet} also suggested to prune filters based on the statistics from the next layer to evaluate a filter's importance.

Quantization reduces the float-number representations of weights and activations to lower numerical precision ~\cite{wu2016quantized, han2015deep}. The extreme case could even just use binary values~\cite{courbariaux2015binaryconnect, rastegari2016xnor}. Beyond scalar quantization, vector quantization was widely adopted too in model compression for parameter sharing~\cite{gong2014compressing,wu2018deep}. 

Recently, \cite{he2018amc,liu2018adadeep} pointed out that conventional model compression techniques rely on domain
experts to explore the large design space trading off, which could be sub-optimal and tedious. The idea of AutoML for model compression has been extended by a number of works, e.g,~\cite{wu2019fbnet, cheng2018instanas,tan2019efficientnet, tan2019mnasnet, cheng2018instanas}, although most of them focused on compressing deep classifiers.


\vspace{-0.5em}
\subsection{GANs and GAN Compression}
\vspace{-0.5em}
GANs have witnessed prevailing success in many practical applications built on image-to-image translation. \textit{CycleGAN} \cite{zhu2017unpaired} and its many successors pioneered in demonstrating GAN's power in image-to-image translation without paired training data. They explored cycle-consistency loss to avoid mode collapse, enabling both diverse and semantically consistent synthesis, therefore making them popular choices in style transfer-type applications. However, GANs with a cycle loss are often costly and difficult to train. Recent works have found that \textit{encoder-decoder GANs}, i.e., fully feedforward generators and discriminators with no cycle structures, can also perform on par, with or without paired training data. These encoder-decoder GANs also gain popularity in image enhancement \cite{wang2018esrgan,kupyn2019deblurgan,jiang2019enlightengan} and editing \cite{sanakoyeu2018style,yang2019controllable}.

Substantial evidences reveal that the generative quality of GANs consistently benefits from larger-scale training \cite{karras2017progressive,brock2018large,gui2020review}. However, the GAN training is notoriously unstable, and therefore numerous techniques were proposed to stabilize the training of increasingly larger GANs, including spectral normalization~\cite{miyato2018spectral}, gradient penalty~\cite{gulrajani2017improved} and progressive training~\cite{karras2017progressive}. 

Despite their performance boost, the growing complexity of those large GANs conflicts the demands of mobile deployments, calling for compression techniques. To our best knowledge, \cite{shu2019co} is the only published method in GAN compression at our submission time. The authors experimentally showed that classical channel pruning would fail when applied to compress GANs, due to the sensitivity of GAN training. They thus proposed a co-evolution algorithm to regularize pruning, that relied on the cycle-consistency loss. As a result, their algorithm was restricted to compressing CycleGAN or its variants, and cannot be applied to encoder-decoder GANs. It also needed the trained discriminator of original GANs as part of the pruning loss.

A concurrent work (available after our submission time) that we shall credit is~\cite{li2020gan}, which also presents AutoML for GAN compression. Its uniqueness lies in searching the channel width for an existing generator. In a ``once-for-all" manner, their searched models achieved impressive image translation performance without fine-tuning. In comparison, our proposed AGD framework customizes the search space for tasks with different properties, searches for operators types in addition to channel widths, and additionally evaluates on a super-resolution task.

\section{Our AutoGAN-Distiller Framework}

Given a pretrained generator $G_0$ over the data $\chi=\{x_i\}^{N}_{i=1}$, our goal is to obtain a more compact and hardware-friendly generator $G$, from $G_0$ while the generated quality does not sacrifice much, e.g., $G_0(x) \approx G(x), x\in \chi$.

We adopt the differentiable design \cite{liu2018darts} for our NAS framework. A NAS framework consists of two key components: the search space and the proxy task.
AGD customizes both these two components for the specific task of GAN compression, as described in the following sections.

\subsection{The Proposed Customized Search Space}\label{sec:search_space}

\textbf{General Design Philosophy.} Many NAS works adopt the directed acyclic graph (DAG)-type search spaces~\cite{liu2018darts} due to its large volume and flexibility. However, those can result in lots of irregular dense connections. As discussed in \cite{ma2018shufflenet}, such network fragmentation is hardware-unfriendly, as it will significantly hamper the parallelism and induce latency/energy overheads. Therefore, we choose a sequential search space, for the ease of implementing the desired parallelism on real mobile devices. In a sequential pipeline, each individual searchable module (a.k.a. a node) only contains a single operator, and they are sequentially connected without any branching. 

We jointly search for the operator type and the width of each layer. Different from previous differentiable search \cite{wu2019fbnet} that selects from predefined building block options where the combinations of operator types and widths are manually picked, we enable each layer's width and operators to be independently searchable.


\textbf{Application-specific Supernet.} On top of the above general design philosophy, we note that different tasks have developed their own preferred and customized components, calling for instantiating different network architectures per application needs. For example, among common image-to-image GAN tasks, image enhancement prefers deeper networks to recover detailed textures and being precise at pixel-level predictions, while style transfer often refers to shallower architectures, and focuses on consistency of the global style more than local details.

We next derive two specific supernet structures below, for two common tasks: image translation and image super-resolution, respectively. Both follow the general sequential search space design with the efficiency constraint in mind.
\begin{itemize}
\vspace{-0.5em}
\item \textit{Image Translation.} Our supernet architecture inherits original CycleGAN's structure (see Fig.~\ref{fig:search_space_st}). We keep the same operators from the stem (the first three convolutional layers) and the header (the last two transposed convolutional layers and one convolutional layer), except searching for their widths. This is motivated by the fact that their downsampling/upsampling structures contribute to lowering the computational complexity of the backbone between them, which have been proved to be effective \cite{zhu2017unpaired}. 

For the backbone part between the stem and header, we adopt nine sequential layers with both searchable operators and widths to trade off between model performance and efficiency. For the normalization layers, we adopt the instance normalization~\cite{ulyanov2016instance} that is widely used in image translation \cite{zhu2017unpaired} and style transfer~\cite{johnson2016perceptual}.
\vspace{-0.2em}

\begin{figure}[h]
\centering
\includegraphics[width=0.48\textwidth]{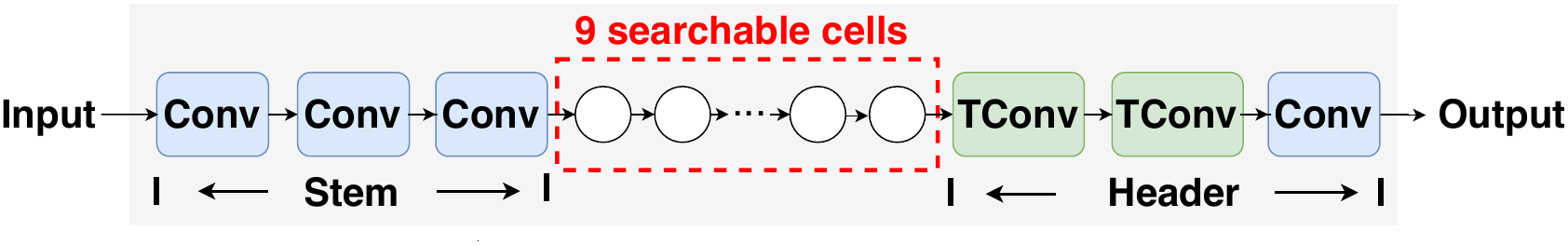}
\vspace{-1.5em}
\caption{Our supernet for Image Translation, where ``TConv'' denotes a transposed convolutional layer. We search for both the operators and widths for the layers inside the red box, leaving the stem and header to only search for widths. }
\vspace{-0.5em}
\label{fig:search_space_st}
\end{figure}

\item \textit{Super Resolution.} Our supernet architecture is inspired by SRResNet~\cite{ledig2017photo} where most computation is performed in the low resolution feature space for efficiency (see Fig.~\ref{fig:search_space_sr}). We fix the stem and header parts in the original design which amount to $<1\%$ computation, and search the remaining residual layers.

For the residual network structure, ESRGAN~\cite{wang2018esrgan} introduces residual-in-residual (RiR) blocks with dense connections which have higher capacity and is easier to train. Despite improved performance, such densely-connected blocks are hardware-unfriendly. Therefore, we replace the dense blocks in residual-in-residual modules with five sequential layers with both searchable operators and widths. We also remove all the batch normalization layers in the network to eliminate artifacts~\cite{wang2018esrgan}. 
\vspace{-0.2em}

\begin{figure}[h]
\centering
\includegraphics[width=0.5\textwidth]{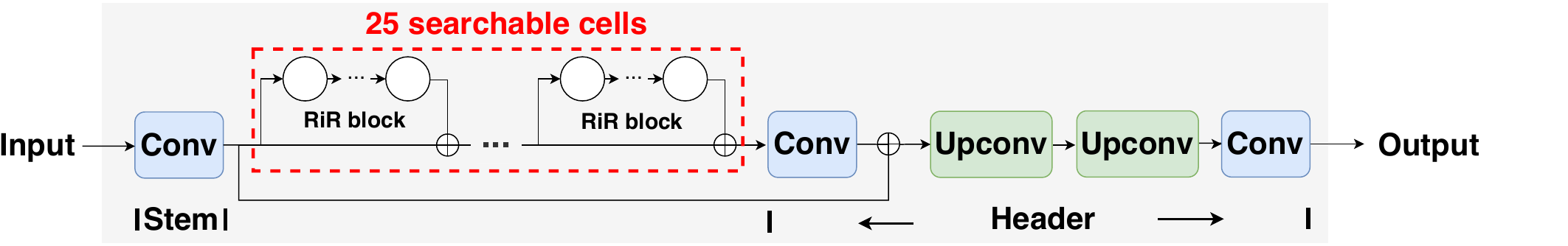}
\vspace{-1.5em}
\caption{Our supernet for Super Resolution, where ``Upconv'' denotes bilinearly upsampling followed by a convolutional layer. We search for both the operators and widths for the layers inside the red box, leaving the stem and header fixed. }
\label{fig:search_space_sr}
\vspace{-0.5em}
\end{figure}

\end{itemize}

\textbf{Operator Search.} We search for the following operators:
\begin{itemize}
\vspace{-0.5em}
    \item Conv 1$\times$1; \quad Conv 3$\times$3; 
    \vspace{-0.3em}
    \item Residual Block (``ResBlock'') (2 layers of Conv 3$\times$3, with a skip connection); 
    \vspace{-0.3em}
    \item Depthwise Block (``DwsBlock'') (Conv1 $\times$1 + DepthwiseConv 3$\times$3 + Conv 1$\times$1, with a skip connection).\vspace{-1em}
\end{itemize}
These cover popular and efficient building blocks that have been used to construct GAN generators. We did not include dilated convolutions, motivated by their hardware unfriendliness as analyzed in~\cite{ma2018shufflenet}.

We search for the operator for each layer in a differentiable manner \cite{liu2018darts,wu2019fbnet}. For the $i$-th layer, we use an architecture parameter $\alpha_i$ to determine the operator for the current layer, and the softmax value of $\alpha_i$ denotes the probability of choosing this operator. All the candidate operators are activated for every iteration in searching process and the output is the weighted sum of all the operators determined by all the softmax values of all $\alpha$.

\textbf{Width Search.} 
 The balance between the capacity of an operator and its width (i.e., number of input/output channels) is a crucial trade-off for a compressed model. Therefore, we also allow for the width search to build a slimmer generator. By searching the widths, we merge the pruning step into the searching process for an end-to-end GAN compression.
 
 However, it is non-trivial to naively build a set of independent convolutional kernels with different widths for each operator, due to the exponentially possible combinations. We thus set a single convolution kernel with a maximal width, named \textit{superkernel}. Then we search for the expansion ratio $\phi$ to make use of only a subset of the input/output dimensions of the superkernel. We set $\phi \in [\frac{1}{3},\frac{1}{2},\frac{3}{4},\frac{5}{6},1]$ and use the architecture parameter $\gamma_i$ to control the probability of choosing each expansion ratio in the $i$-th layer. We apply Gumbel-Softmax~\cite{gumbel1948statistical} to approximate differentiable sampling for $\phi$ based on $\gamma$ as~\cite{cai2018proxylessnas}. Therefore, during the searching process only one most likely expansion ratio will be activated each time, which saves both memory and computation cost.






\subsection{The Proposed Customized Proxy Task}\label{sec:loss_design}

During the search process, the NAS framework is guided by the proxy task to optimize the architecture towards the desired patterns. As our objective is to obtain both superior performance and extreme efficiency, we need to customize our proxy task to supervise our search process in a proper way. We formulate the training objective of our AGD framework in Eq.~\ref{eqn:objective}:
\begin{equation}
    \min \limits_{G,\alpha,\gamma} \frac{1}{N} \sum_{i=1}^{N} d(G(x_i, \alpha, \gamma),G_0(x_i))+\lambda F(\alpha, \gamma).
\label{eqn:objective}
\end{equation}
Here, $d(\cdot, \cdot)$ is a distance metric for the knowledge distillation~\cite{polino2018model} between the compact generator $G$ and the pretrained one $G_0$, $F$ is the computational budget determined by the network architecture,  $\alpha$ and $\gamma$ are the architecture parameters controlling the operator and width of each layer, respectively, and $\lambda$ is the trade-off parameter. Note that both $d(\cdot, \cdot)$ and $F$ are functions of $\alpha$ and $\gamma$.

AGD searches for an efficient generator $G$ under the guidance of distillation from the original model $G_0$, through optimizing Eq.~\ref{eqn:objective}. 
Note that, the objective function in Eq.~\ref{eqn:objective} is free of any trained discriminator. We choose so because: (1) in practice, the discriminator is often discarded after the generator is trained, therefore not necessarily available when compressing the generator in future; and (2) we also tried to add a discriminator loss term into Eq.~\ref{eqn:objective}, similarly to \cite{shu2019co} did, but found no improvement in this way (actually, the search becomes more unstable). 


One extra hassle arises from our decoupled search of operators and widths. During the search process, if we jointly update $\alpha$ and $\gamma$, we observe that our model is prone to suffering from the same ``architecture collapse'' problem observed in \cite{chen2019fasterseg}, i.e., NAS is quickly biased towards some low-latency yet low-performance models. Therefore, we decouple the computational budget calculation into two individual terms, i.e., the operator-aware and width-aware parts, and weight the two differently:
\begin{equation}
    F(\alpha,\gamma) =  \omega_1 F(\alpha|\gamma) + \omega_2 F(\gamma|\alpha)
\label{eqn:calculate_F}
\end{equation}
In order for more flexible model space exploration while enforcing the budget, we set an upper bound and a lower bound for the target computational budget, and double (half) $\lambda$ when the computational budget of the derived architecture determined by the current architecture parameters is larger (smaller) than the upper (lower) bound.

The last remaining pieces of our framework are the specific choices of $d(\cdot, \cdot)$ and $F$ in Eq.~\ref{eqn:objective}. For $d(\cdot, \cdot)$, we use a combination of content loss $L_{c}$ (avoid color shift) and perceptual loss $L_{p}$ (preserve visual and semantic details) as defined in~\cite{johnson2016perceptual}, in addition to a total variation loss $L_{tv}$ (enforce pixel-level similarity)~\cite{aly2005image}:
\begin{equation}
    d(G,G_0) = \beta_1L_{c}(G, G_0) + \beta_2 L_{p}(G, G_0) + \beta_3 L_{tv}(G, G_0),
\label{eqn:loss}
\end{equation} 
where $\beta_1,\beta_2,\beta_3$ are hyperparameters. For $F$, we apply FLOPs as the computational budget in Eq.~\ref{eqn:objective}.

We summarize our AGD framework in Algorithm~\ref{alg:AGD}. 



\begin{figure}[t]
\begin{minipage}{0.48\textwidth}
\vspace{-0.8em}
\begin{algorithm}[H]
  \caption{The Proposed AutoGAN-Distiller Framework}
  \label{alg:AGD}
\begin{algorithmic}[1]
  \STATE \textbf{Input:} dataset $\chi=\{x_i\}^{N}_{i=1}$, pretrained generator $G_0$, search space and supernet $G$, epochs to pretrain ($T_1$), search ($T_2$) and train-from-scratch ($T_3$)
  \STATE \textbf{Output:} trained efficient generator $G^*$
  \STATE Equally split $\chi$ into $\chi_1$ and $\chi_2$ and initialize supernet weight $w$ and architecture parameters $\{\alpha,\gamma\}$ 
  with uniform distribution 
  \STATE \textit{\# First Step: Pretrain}
  \FOR{$t$ $\gets$ 1\ to\ $T_1$ } 
    \STATE Get a batch of data $X_1$ from $\chi_1$
      \FOR{$\gamma$ in [$\gamma_{\mathrm{max}}$, $\gamma_{\mathrm{min}}$, $\gamma_{\mathrm{random1}}$, $\gamma_{\mathrm{random2}}$]\footnotemark}
        \STATE  $g_w^{(t)} = \nabla_w d(G(X_1, \alpha, \gamma),G_0(X_1))$
        \STATE  $w^{(t+1)}$ = update($w^{(t)}$, $g_w^{(t)}$)
      \ENDFOR
  \ENDFOR
   
  \STATE \textit{\# Second Step: Search}
    \FOR{$t$ $\gets$ 1 \ to\ $T_2$ }
    \STATE Get a batch of data $X_1$ from $\chi_1$
    \STATE  $g_w^{(t)} = \nabla_w d(G(X_1, \alpha, \gamma),G_0(X_1))$
    \STATE  $w^{(t+1)}$ = update($w^{(t)}$, $g_w^{(t)}$)
    \STATE Get a batch of data $X_2$ from $\chi_2$
    \STATE  {\small $g_{\alpha}^{(t)} = \nabla_{\alpha} d(G(X_2, \alpha, \gamma),G_0(X_2)) + \lambda \cdot \omega_1 \nabla_{\alpha} F(\alpha|\gamma) $}
    \STATE  {\small $g_{\gamma}^{(t)} = \nabla_{\gamma} d(G(X_2, \alpha, \gamma),G_0(X_2)) + \lambda \cdot \omega_2 \nabla_{\gamma} F(\gamma|\alpha) $}
    \STATE  $\alpha^{(t+1)}$ = update($\alpha^{(t)}$, $g_{\alpha}^{(t)}$)
    \STATE  $\gamma^{(t+1)}$ = update($\gamma^{(t)}$, $g_{\gamma}^{(t)}$)
  \ENDFOR
  
  \STATE \textit{\# Third Step: Train from scratch}
  \STATE Derive the searched architecture $G^*$ with maximal $\{\alpha,\gamma\}$ for each layer and re-initialize weight $w$.
    \FOR{$t$ $\gets$ 1\ to\ $T_3$ }
    \STATE Get a batch of data $X$ from $\chi$
      \STATE  $g_w^{(t)} = \nabla_w d(G^*(X),G_0(X))$
      \STATE  $w^{(t+1)}$ = update($w^{(t)}$, $g_w^{(t)}$)
  \ENDFOR
\end{algorithmic}
\end{algorithm} 
\end{minipage}
\vspace{-2em}
\end{figure}
\footnotetext{\footnotesize We train each operator with the largest width, the smallest width and two random widths during pretraining, following the ``sandwich rule" in~\cite{yu2019universally}.}
\section{Experiment Results}
\label{sec:experiment}
\subsection{Considered Tasks \& Models.} 
\textbf{Unpaired Image-to-image Translation.} We apply AGD on compressing CycleGAN~\cite{zhu2017unpaired} and consider two datasets, horse2zebra~\cite{zhu2017unpaired} and summer2winter~\cite{zhu2017unpaired}. In particular, we individually conduct the architecture search for each task in one dataset. We also consider a special case for AGD that all the weights and activations are quantized to 8-bit integers for hardware-friendly implementation.

\textbf{Super Resolution.} We apply AGD on compressing ESRGAN~\cite{wang2018esrgan} on a combined dataset of DIV2K and Flickr2K~\cite{Agustsson_2017_CVPR_Workshops} with a upscale factor of 4$\times$, following \cite{wang2018esrgan}. We evaluate the searched/compressed model on several popular SR benchmarks, including Set5~\cite{bevilacqua2012low}, Set14~\cite{zeyde2010single},  BSD100~\cite{martin2001database} and Urban100~\cite{huang2015single}.

\subsection{Evaluation Metrics.} Both tasks are mainly evaluated by visual quality. We also use FID (Frechet Inception Distance)~\cite{heusel2017gans} for the unpaired image-to-image translation task, and PSNR for super resolution, as quantitative metrics. 

For the efficiency aspect, we measure the model size and the inference FLOPs (floating-point operations). 
As both might not always be aligned with the hardware performance, we further measure the real-device inference latency using NVIDIA GEFORCE RTX 2080 Ti~\cite{2080ti}.


\subsection{Training details.} 
\textbf{NAS Search.} The AGD framework adopts the differential search algorithm \cite{liu2018darts}. We split the training dataset into two halves: one for updating supernet weight and the other for updating architecture parameters. The entire search procedure consists of three steps:
\begin{itemize}
    \vspace{-1em}
    \item Pretrain. We only train the supernet weights on the first half dataset. To adapt the supernet for different expansion ratios, we train each operator with the largest width, the smallest width, and two random widths, following the ``sandwich rule" in~\cite{yu2019universally}.
    \vspace{-1.8em}
    \item Search. We then search the architecture by alternatively updating supernet weights and the architecture parameters $\{\alpha, \gamma\}$. During the training, the width of each layer is sampled through Gumbel-Softmax based on $\gamma$ and the output is the weighted sum of the results of all operators based on the softmax value of $\alpha$.
    \vspace{-0.5em}
    \item Derive. We derive the final architecture by choosing the operator and width with the maximal probability for each layer, determined by $\alpha$ and $\gamma$. We then train the derived architecture from scratch.
    \vspace{-0.9em}
\end{itemize}


\textbf{Unpaired Image-to-image Translation.} For AGD on CycleGAN, $\lambda$ in Eq.~\ref{eqn:objective} is $1\times10^{-17}$, $\omega_1$ and $\omega_2$ in Eq.~\ref{eqn:calculate_F} are set to 1/4 and 3/4, and $\beta_1,\beta_2$ and $\beta_3$ in Eq.~\ref{eqn:loss} are set to be $1\times10^{-2}$, 1, and $5\times10^{-8}$, respectively. We pretrain and search for 50 epochs, with batch size 2. We use an SGD optimizer with a momentum of 0.9 and the initial learning rate $1\times 10^{-1}$ for the weights, which linearly decays to 0 after 10 epochs, and an Adam optimizer with a constant $3\times 10^{-4}$ learning rate for architecture parameters. We train the searched architecture from scratch for 400 epochs, with a batch size of 16 and an initial learning rate of $1\times 10^{-1}$, which linearly decays to 0 after 100 epochs. 

\textbf{Super Resolution.} For AGD on ESRGAN, $\lambda$ in Eq.~\ref{eqn:objective} is $1\times10^{-12}$, $\omega_1$ and $\omega_2$ in Eq.~\ref{eqn:calculate_F} are 2/7 and 5/7, and $\beta_1,\beta_2$, and $\beta_3$ in Eq.~\ref{eqn:loss} are set to be $1\times 10^{-2}$, 1, and $5\times 10^{-8}$, respectively. We pretrain and search for 100 epochs, with each batch of 6 cropped patches with $32\times32$.
We set the initial learning rate for the weights to be $1\times 10^{-4}$ with an Adam optimizer, decayed by 0.5 at the 25-th, 50-th, and 75-th epoch, and that for the architecture parameters is a constant of $3\times 10^{-4}$. 
To train the searched architecture from scratch, we 
train for 1800 epochs with a learning rate of $1\times 10^{-4}$, decayed by 0.5 at 225-th, 450-th, 900-th, and 1300-th epoch, following ~\cite{wang2018esrgan}.


\begin{table}[htbp]
  \centering
  \vspace{-0.7em}
  \caption{Statistics of the original CycleGAN~\cite{zhu2017unpaired} on different unpaired image translation tasks, where the latency is measured on NVIDIA GEFORCE RTX 2080 Ti. Note that the FID of CycleGAN is from CEC~\cite{shu2019co} since FID~\cite{heusel2017gans} is proposed after CycleGAN~\cite{zhu2017unpaired}.}
  \vspace{0.5em}
    \begin{tabular}{c|cc|c}
    \toprule
    \multicolumn{1}{c|}{GFLOPs} & \multicolumn{1}{c|}{54.17} & \multicolumn{1}{c|}{Latency (ms)} & 7.25 \\
    \midrule
    \multicolumn{1}{c|}{Memory (MB)} & \multicolumn{3}{c}{43.51} \\
    \midrule
    \multicolumn{1}{c|}{\multirow{4}[6]{*}{FID}} & \multicolumn{2}{c|}{horse2zebra} & 74.04 \\
    \cmidrule{2-4}       & \multicolumn{2}{c|}{zebra2horse} & 148.81 \\
    \cmidrule{2-4}       & \multicolumn{2}{c|}{summer2winter} & 79.12 \\
    \cmidrule{2-4}       & \multicolumn{2}{c|}{winter2summer} & 73.31 \\
    \bottomrule
    \end{tabular}%
  \label{table:orig_cyclegan}%
  \vspace{-1em}
\end{table}%

\subsection{AGD for Efficient Unpaired Image Translation}
For unpaired image  translation, the original performance and efficiency statistics of CycleGAN are reported in Table~\ref{table:orig_cyclegan}. We compare our AGD with CEC~\cite{shu2019co}, the existing GAN compression algorithm specifically for CycleGAN; we also include classical structural pruning \cite{li2016pruning} developed for classification models\footnote{We prune 70\% filters with smallest $\ell_1$-norm values in each layer, and then finetune the remaining weights.}.



\textbf{Quantitative Results.} Table~\ref{table:exp_st} demonstrates that for all tasks, AGD consistently achieves significantly better FID than the state-of-the-art CEC, with even higher savings of model sizes and FLOPs. Besides, the performance of structural pruning, which is effective for compressing classification models, lags far behind with CEC, not to say AGD. That re-confirms \cite{shu2019co}'s finding that existing classification-oriented compression algorithms cannot be directly plugged into compressing GANs.


Furthermore, the gain of AGD also manifests in real-device latency measurements. For example, on the summer2winter dataset, the per-image inference latency of the AGD-compressed model is reduced by 59.86\% with an even slightly better FID than the original CycleGAN.


\begin{table}[htbp]
  \centering
  \vspace{-1em}
  \caption{Quantitative comparison with the structural pruning~\cite{li2016pruning} and CEC~\cite{shu2019co} on CycleGAN compression in different unpaired image translation tasks.}
  \vspace{0.5em}
  \resizebox{0.48\textwidth}{!}{
    \begin{tabular}{cccccc}
    \toprule
    \multicolumn{1}{c}{Dataset} & Method & \multicolumn{1}{c}{FID} & \multicolumn{1}{c}{GFlops} & \multicolumn{1}{c}{Memory (MB)} & \multicolumn{1}{c}{Latency (ms)} \\
    \midrule
    \midrule
    \multicolumn{1}{c}{\multirow{3}[6]{*}{horse2zebra}} & Prune  & 220.91 & 4.95   & 3.91   & 6.19 \\
    \cmidrule{2-6}       & CEC    & 96.15  & 13.45  & 10.16  & \multicolumn{1}{c}{-} \\
    \cmidrule{2-6}       & \textbf{AGD} & \textbf{83.6} & \textbf{6.39} & \textbf{4.48} & \textbf{4.07} \\
    \midrule
    \multicolumn{1}{c}{\multirow{3}[5]{*}{zebra2horse}} & Prune  & 206.56 & 4.95   & 3.91   & 6.19 \\
    \cmidrule{2-6}       & CEC    & 157.9  & 13.06  & 10     & \multicolumn{1}{c}{-} \\
    \cmidrule{2-6}       & \textbf{AGD} & \textbf{137.2} & \textbf{4.84} & \textbf{3.2} & \textbf{3.99} \\
    \midrule
    \multicolumn{1}{c}{\multirow{3}[5]{*}{summer2winter}} & Prune  & 120.94 & 4.95   & 3.91   & 6.19 \\
    \cmidrule{2-6}       & CEC    & 78.58  & 12.98  & 7.98   & \multicolumn{1}{c}{-} \\
    \cmidrule{2-6}       & \textbf{AGD} & \textbf{78.33} & \textbf{4.34} & \textbf{2.72} & \textbf{2.91} \\
    \midrule
    \multicolumn{1}{c}{\multirow{3}[6]{*}{winter2summer}} & Prune  & 106.2  & 4.95   & 3.91   & 6.19 \\
    \cmidrule{2-6}       & CEC    & 79.26  & 16.45  & 7.61   & \multicolumn{1}{c}{-} \\
    \cmidrule{2-6}       & \textbf{AGD} & \textbf{77.73} & \textbf{4.26} & \textbf{2.64} & \textbf{4.26} \\
    \bottomrule
    \end{tabular}%
    }
    \vspace{-0.5em}
  \label{table:exp_st}%
\end{table}%

\begin{figure*}[!th] 
	\centering
	\setlength{\tabcolsep}{3pt}
	\begin{tabular}{cccc}
		Source images & Original CycleGAN & CEC \cite{shu2019co} & AGD (Proposed)
		\\
		& \makecell{54.17 GFLOPs\\43.51 MB} & \makecell{13.45 GFLOPs\\10.16 MB} & \makecell{6.39 GFLOPs\\4.48 MB} \\
        \includegraphics[width=0.22\linewidth]{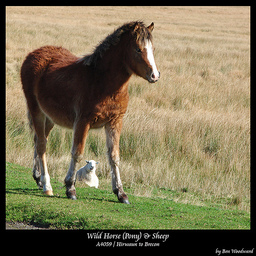} & 
		\includegraphics[width=0.22\linewidth]{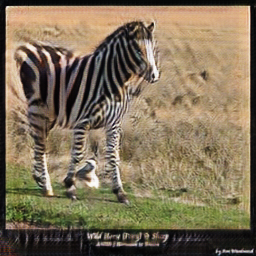} & 
		\includegraphics[width=0.22\linewidth]{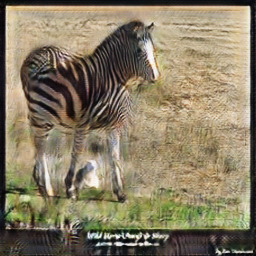} &
		\includegraphics[width=0.22\linewidth]{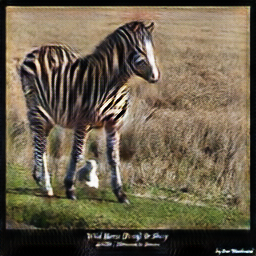} 
		\\
        \includegraphics[width=0.22\linewidth]{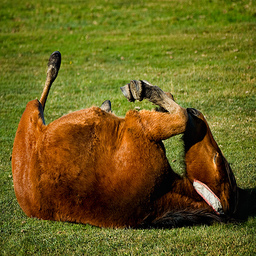} & 
		\includegraphics[width=0.22\linewidth]{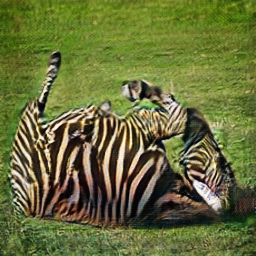} & 
		\includegraphics[width=0.22\linewidth]{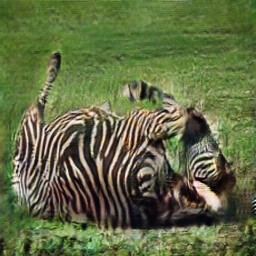} &
		\includegraphics[width=0.22\linewidth]{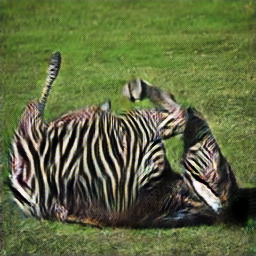}  
		\\
		& \makecell{54.17 GFLOPs\\43.51 MB} & \makecell{13.06 GFLOPs\\10.00 MB} & \makecell{4.84 GFLOPs\\3.20 MB} \\
        \includegraphics[width=0.22\linewidth]{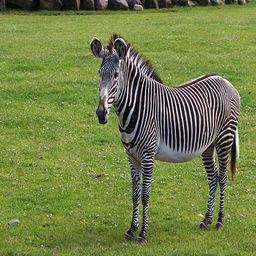} & 
        \includegraphics[width=0.22\linewidth]{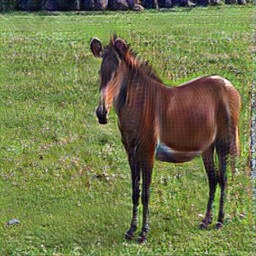} & 
        \includegraphics[width=0.22\linewidth]{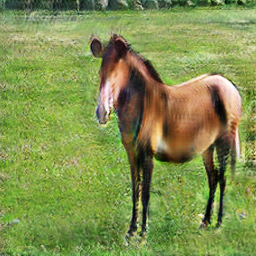} & 
        \includegraphics[width=0.22\linewidth]{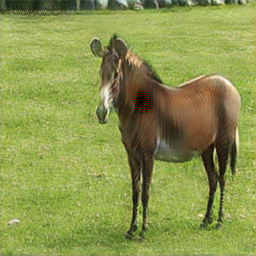} 
		\\
		& \makecell{54.17 GFLOPs\\43.51 MB} & \makecell{12.98 GFLOPs\\7.98 MB} & \makecell{4.34 GFLOPs\\2.72 MB} \\
        \includegraphics[width=0.22\linewidth]{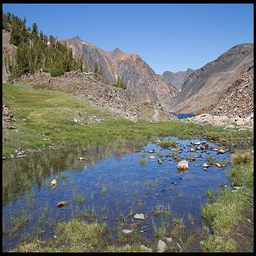} & 
        \includegraphics[width=0.22\linewidth]{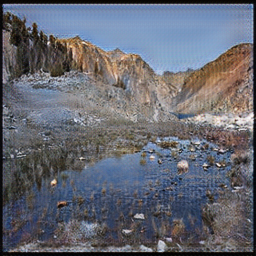} & 
        \includegraphics[width=0.22\linewidth]{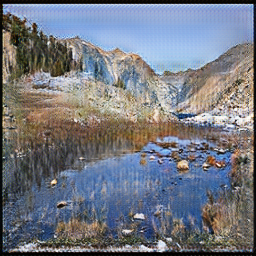} & 
        \includegraphics[width=0.22\linewidth]{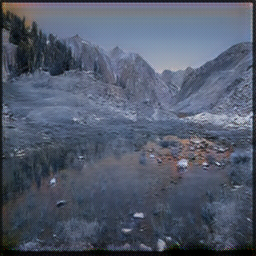} 
	\end{tabular}
		\vspace{-0.5em}
	\caption{Visualization examples of CycleGAN compression on the horse2zebra task (rows 1 and 2), zebra2horse task (row 3), and summer2winter task (row 4). Columns from left to right: source images, translation results by original CycleGAN, CEC, and AGD, respectively, and the FLOPs and memory (model size) of each method on each task are annotated above the images.}
	\label{fig:visual_st}
	\vspace{-1.0em}
\end{figure*}

\textbf{Visualization Results.} We compare the visual quality from the original CycleGAN, CEC, and AGD in Fig.~\ref{fig:visual_st}, using examples from the \textit{horse2zebra} and \textit{zebra2horse} tasks. The AGD-compressed CycleGAN yields comparable visual quality to the uncompressed model in all cases. With zooming-in, AGD also appears to visually outperform CEC, in producing sharper edges and textural details, suppressing checkboard artifacts, as well as eliminating color distortion. 

\begin{table}[!h]
  \centering
  \vspace{-0.5em}
  \caption{The searched architecture (operator, width) by our AGD framework on CycleGAN.}
  \vspace{0.5em}
  \resizebox{0.48\textwidth}{!}{
    \begin{tabular}{c|ccccc}
    \hline
    Block ID & Stem0 & Stem1 & Stem2 & B1 & B2 \\ \hline
    (OP, Width) & (-, 88) & (-, 88) & (-, 88) & (DwsBlock, 88) & (ResBlock, 88) \\ \hline
    Block ID & B3 & B4 & B5 & B6 & B7 \\ \hline
    (OP, Width) & (ResBlock, 88) & (ResBlock, 128) & (ResBlock, 88) & (DwsBlock, 128) & (Conv1x1, 216) \\ \hline
    Block ID & B8 & B9 & Header1 & Hearder2 & Header3 \\ \hline
    (OP, Width) & (Conv1x1, 88) & (Conv1x1, 128) & (-, 216) & (-, 88) & (-, 3) \\ \hline
    \end{tabular}
    }
    
  \label{table:cyclegan_arch}%
\end{table}%

\vspace{-0.5em}
\textbf{AGD Searched Architecture.}
The searched architecture by AGD on the horse2zebra dataset is outlined in Table~\ref{table:cyclegan_arch}. AGD selects the ``heavier'' DwsBlocks and ResBlocks with narrower channel widths for the early layers; while leaving the last three layers to be Conv 1x1, with larger widths. The searched design remains to be overall light-weighted, while not ``collapsing'' to overly under-parameterized, poor performance models. Similar architectures searched on the other three tasks are shown in our supplementary materials.

\begin{figure}[!t] 
	\centering
	\setlength{\tabcolsep}{1pt}
	\begin{tabular}{ccc}
        \includegraphics[width=0.3\linewidth]{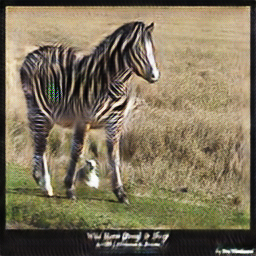} & 
        \includegraphics[width=0.3\linewidth]{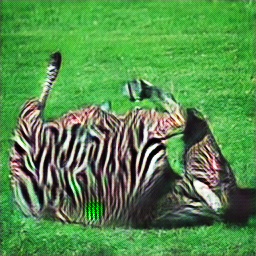} & 
        \includegraphics[width=0.3\linewidth]{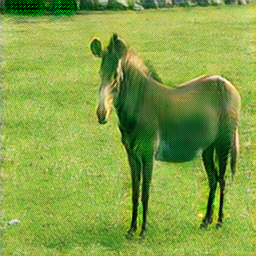} 
	\end{tabular}
	\vspace{-0.5em}
	\caption{Visualization results of CycleGAN compressed by AGD with quantization, where the first two figures are from the \textit{horse2zebra} task,and the last figure is from the \textit{zebra2horse} task.}
	\label{fig:visual_st_quant}
	\vspace{-1em}
\end{figure}

\textbf{Quantization.} We also conduct an extended study on how our AGD compression work with quantization. On the AGD-compressed CycleGAN, we further quantize all the weights and activations to 8-bit integers, following~\cite{banner2018scalable}. The quantitative and visualization results are shown in Table~\ref{table:exp_quant} and Fig.~\ref{fig:visual_st_quant}, respectively: after being further aggressively compressed, the models maintain both competitive FID and comparable visualization quality. 

\begin{table}[htbp!]
  \centering
  \vspace{-1em}
  \caption{Performance of AGD framework on CycleGAN compression with 8-bit quantization.}
  \vspace{0.5em}
    \begin{tabular}{cccc}
    \toprule
    Datasets  & FID    & Memory (MB) \\
    \midrule
    \midrule
    horse2zebra & 85.74   & 1.00 \\
    zebra2horse  & 140.08   & 1.24 \\
    \bottomrule
    \end{tabular}%
  \label{table:exp_quant}%
  \vspace{-1em}
\end{table}%

\begin{figure*}[th!] 
	\centering
	\setlength{\tabcolsep}{1pt}
	\begin{tabular}{ccccc}
	Original ESRGAN & Pruned ESRGAN & SRGAN & VDSR & AGD (Proposed) \\
	\makecell{1176.61 GFLOPs \\ 66.8 MB} & \makecell{113.07 GFLOPs \\ 6.40 MB} & \makecell{166.66 GFLOPs \\ 6.08 MB} & \makecell{699.36 GFLOPs \\ 2.67 MB} & \makecell{108.06 GFLOPs \\ 1.64 MB} \\
        \includegraphics[width=0.19\linewidth]{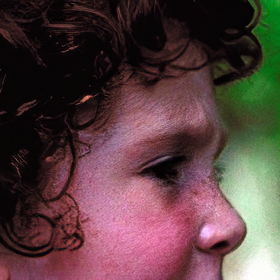} & 
        \includegraphics[width=0.19\linewidth]{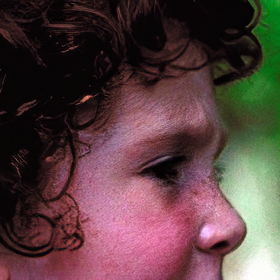} & 
        \includegraphics[width=0.19\linewidth]{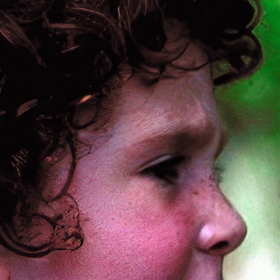} & 
        \includegraphics[width=0.19\linewidth]{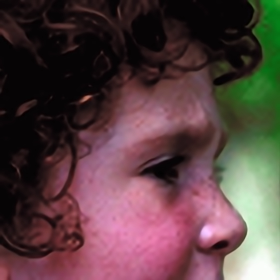} & 
        \includegraphics[width=0.19\linewidth]{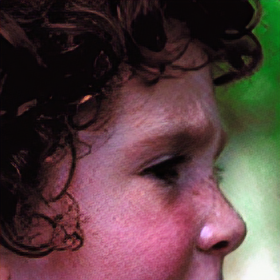} 
		\\
		
        \includegraphics[width=0.19\linewidth]{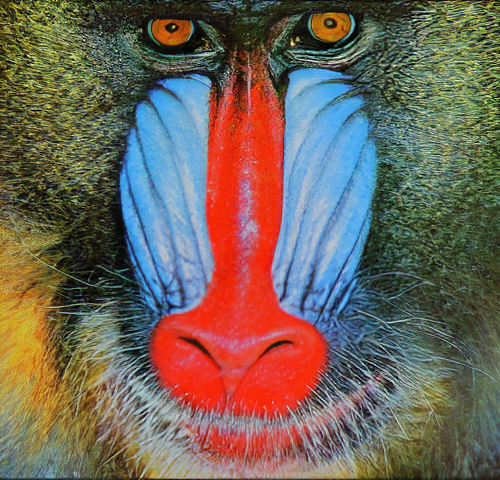} & 
        \includegraphics[width=0.19\linewidth]{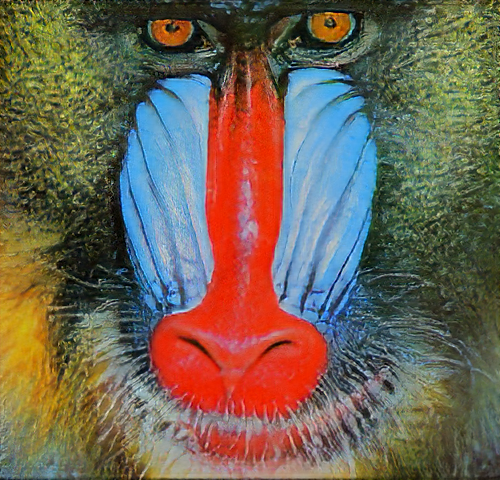} & 
        \includegraphics[width=0.19\linewidth]{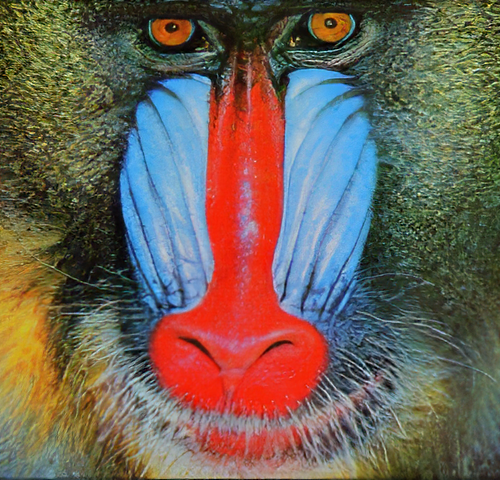} & 
        \includegraphics[width=0.19\linewidth]{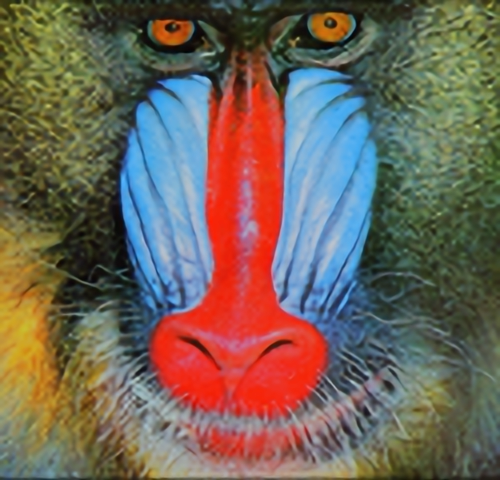} & 
        \includegraphics[width=0.19\linewidth]{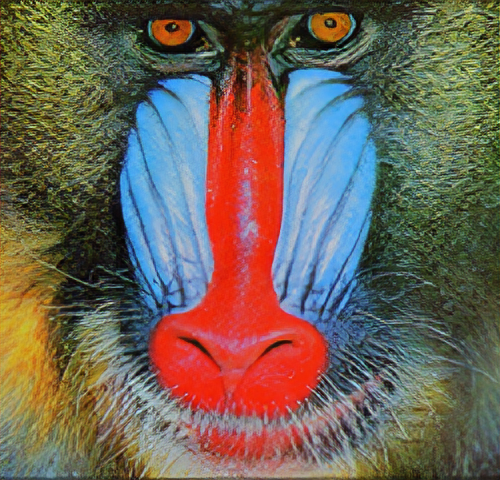} 
		\\
		
        \includegraphics[width=0.19\linewidth]{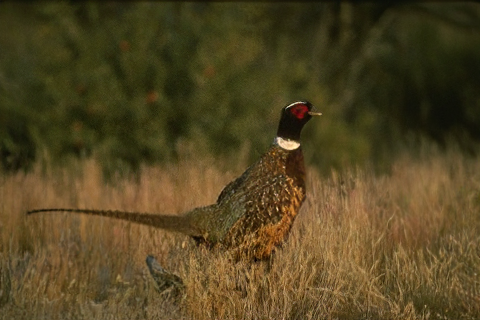} & 
        \includegraphics[width=0.19\linewidth]{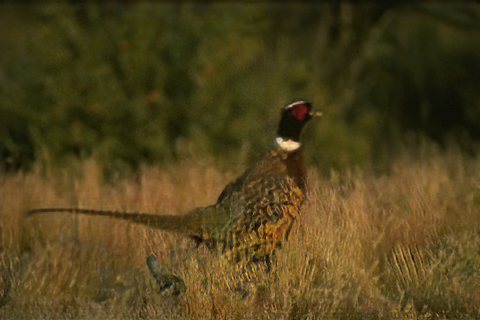} & 
        \includegraphics[width=0.19\linewidth]{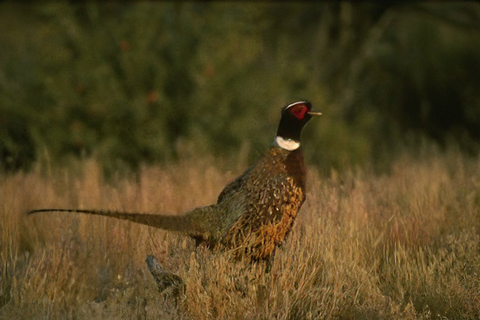} & 
        \includegraphics[width=0.19\linewidth]{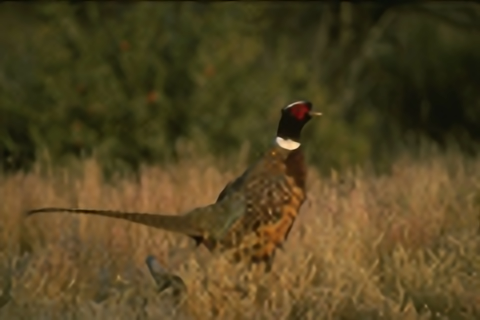} & 
        \includegraphics[width=0.19\linewidth]{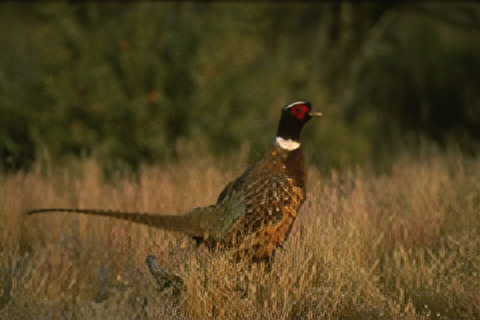} 
		\\

		 \includegraphics[width=0.19\linewidth]{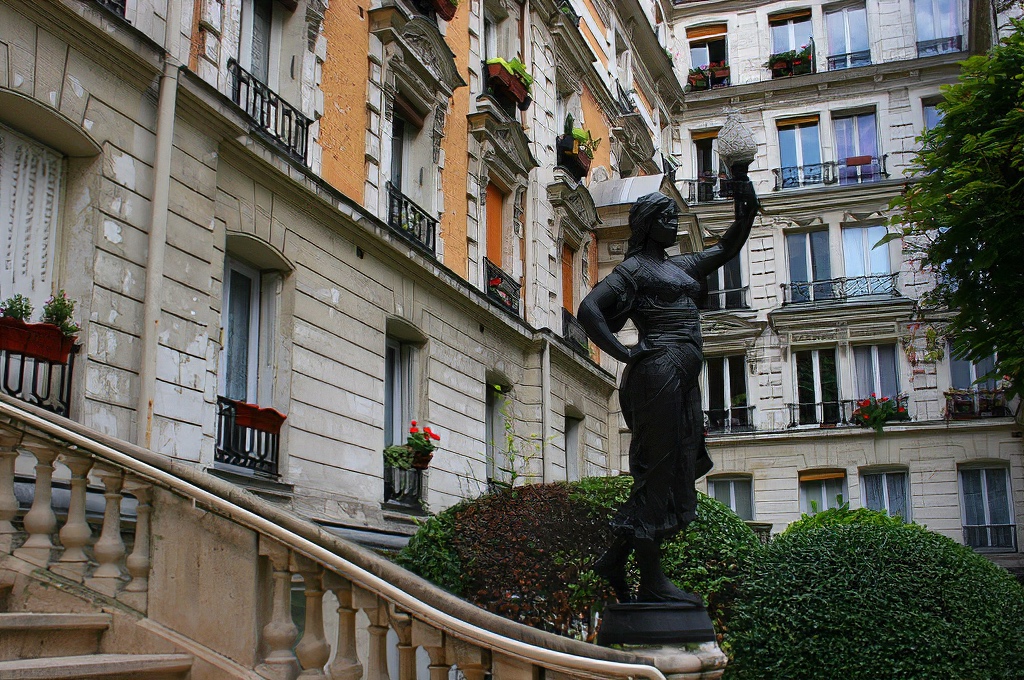} & 
        \includegraphics[width=0.19\linewidth]{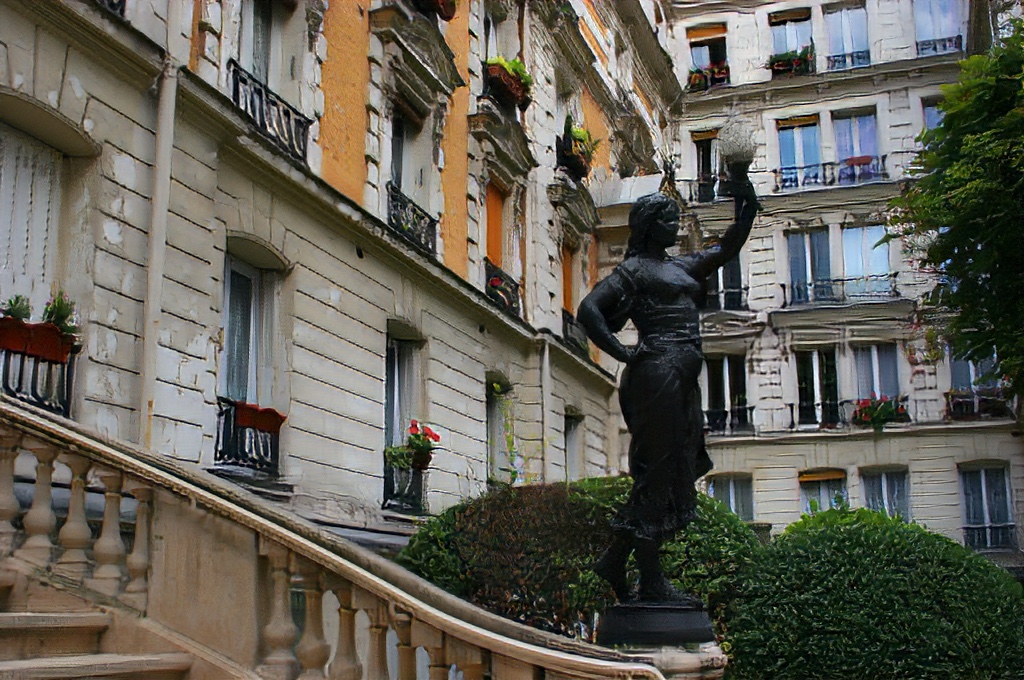} & 
        \includegraphics[width=0.19\linewidth]{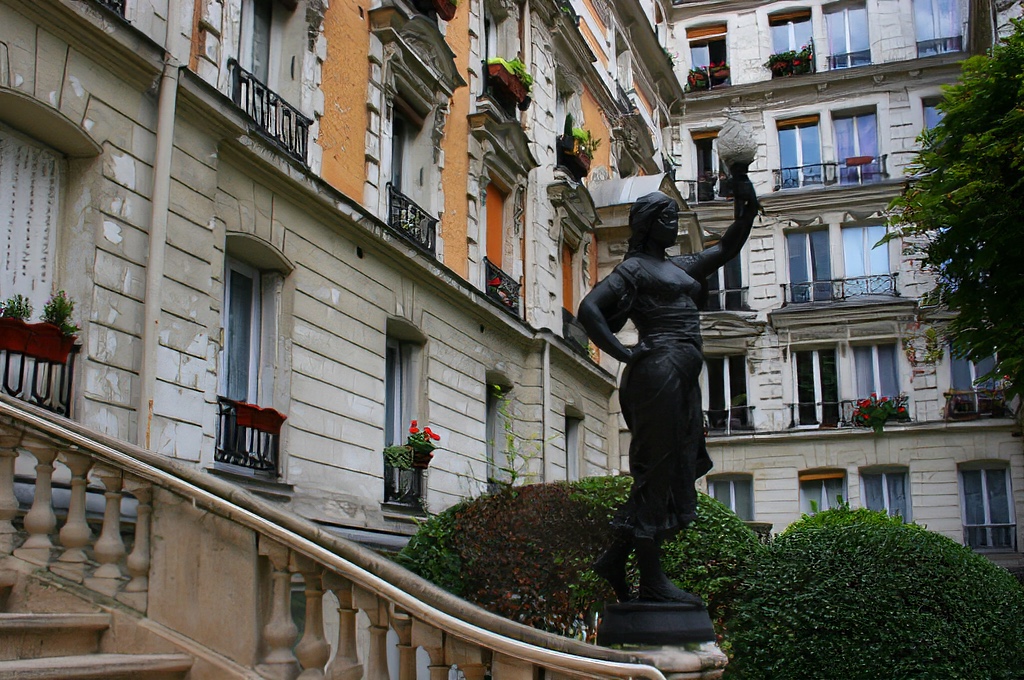} & 
        \includegraphics[width=0.19\linewidth]{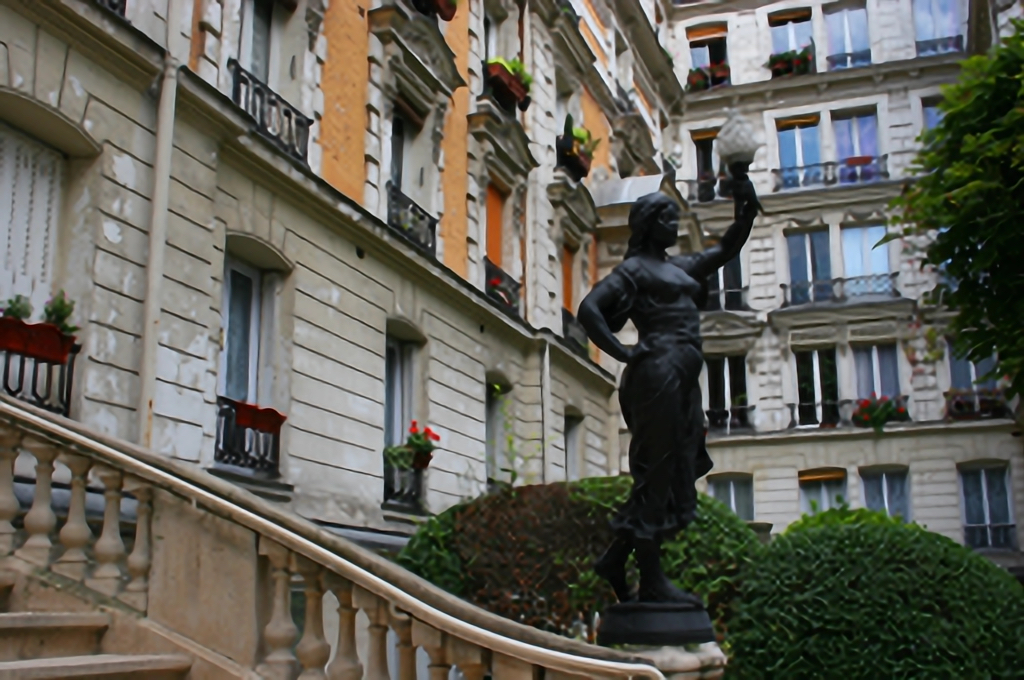} &
        \includegraphics[width=0.19\linewidth]{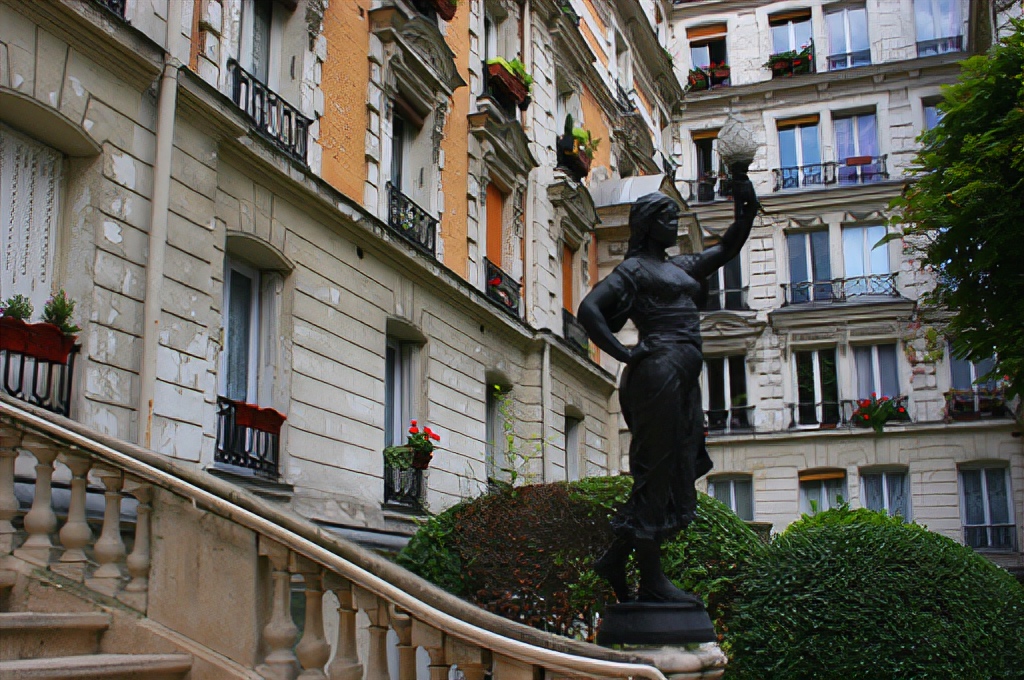} 
		\\
	\end{tabular}
	\vspace{-1.0em}
	\caption{Visualization results of different super resolution methods (Better viewed after zooming in). From top to bottom rows are examples from Set5, Set14, BSD100, and Urban100. Columns from left to right: original ESRGAN~\cite{wang2018esrgan}, ESRGAN after structural pruning \cite{li2016pruning}, SRGAN~\cite{ledig2017photo}, VDSR~\cite{kim2016accurate}, and ESRGAN compressed by AGD. FLOPs (calculated as processing a 256$\times$256 low resolution image with a scale factor of four) and memory (model size) of each method on each task are annotated above the images.}
	\label{fig:visual_sr}
	\vspace{-1.5em}
\end{figure*}

\subsection{AGD for Efficient Super Resolution}
We then apply AGD on compressing ESRGAN~\cite{wang2018esrgan}, a start-of-the-art GAN for super resolution. ESRGAN does not adopt a CycleGAN-type structure (its generator is a feedforward encoder-decoder), therefore CEC \cite{shu2019co} is not applicable. We compare AGD with a structural pruning~\cite{li2016pruning} baseline, and also include two other high-performance and lighter SR models: SRGAN~\cite{ledig2017photo} and VDSR~\cite{kim2016accurate}, to better evaluate the utility-efficiency trade-off of our compressed models.
Besides, since visual quality is the ultimate demand in SR, we first adopt the visualization-oriented ESRGAN\footnote{ESRGAN~\cite{wang2018esrgan} provides two pretrained models, including the PSNR-oriented one trained only with the content loss, targeted for higher PSNR performance; and the visualization-oriented one trained with both content loss and perceptual loss, targeted for higher perceptual quality.} as the teacher model in Eq.~\ref{eqn:loss}.



\begin{table}[!t]
  \vspace{-0.5em}
  \centering
  \caption{Quantitative comparison with the state-of-the-art GAN-based visualization-oriented SR models, where the FLOPs are calculated as processing a 256$\times$256 low resolution image with a scale factor of four.}
  \vspace{0.5em}
  \resizebox{0.48\textwidth}{!}{
    \begin{tabular}{cccccc}
    \toprule
    \multicolumn{2}{c}{Model} & \multicolumn{1}{c}{ESRGAN} & \multicolumn{1}{c}{ESRGAN-Prune} & \multicolumn{1}{c}{SRGAN} & \multicolumn{1}{c}{AGD} \\
    \midrule
    \multicolumn{2}{c}{GFLOPs (256x256)} & 1176.61 & 113.07 & 166.66 & \textbf{108.6} \\
    \midrule
    \multicolumn{2}{c}{Memory (MB)} & 66.8   & 6.40    & 6.08   & \textbf{1.64} \\
    \midrule
    \midrule
    \multicolumn{1}{c}{\multirow{2}[4]{*}{Set5}} & PSNR   & \textbf{30.47}  &   28.07     & 29.40   & \textbf{30.44} \\
    \cmidrule{2-6}      & Latency (ms) & 84.62  & 64.75  & \textbf{4.92}   & \textbf{5.08} \\
    \midrule
    \multicolumn{1}{c}{\multirow{2}[4]{*}{Set14}} & PSNR   & 26.29  &  25.21   & 26.02  & \textbf{27.28} \\
    \cmidrule{2-6}        & Latency (ms) & 80.03  & 78.47  & \textbf{5.37}   & \textbf{6.91} \\
    \midrule
    \multicolumn{1}{c}{\multirow{2}[4]{*}{BSD100}} & PSNR   & 25.32  & 24.74  & 25.16  & \textbf{26.23} \\
   \cmidrule{2-6}       & Latency (ms) & 81.34  & 69.43  & \textbf{4.15}   & \textbf{5.11} \\
    \midrule
    \multicolumn{1}{c}{\multirow{2}[4]{*}{Urban100}} & PSNR   & 24.36  & 22.67  & 24.39  & \textbf{24.74} \\
    \cmidrule{2-6}        & Latency (ms) & 184.79 & 113.82 & 25.28  & \textbf{21.60} \\
    \bottomrule
    \end{tabular}%
    }
  \label{table:exp_sr}%
  \vspace{-1em}
\end{table}%

\textbf{Quantitative Results.}  Table~\ref{table:exp_sr} shows the quantitative comparison in PSNR of a series of GAN-based super resolution methods, re-confirming the little-to-no performance loss of AGD on ESRGAN (interestingly, sometimes even PSNR improvement is seen). Compared to the off-the-shelf alternatives (e.g., SRGAN), AGD compressed ESRGAN obtains consistently superior PSNRs, with a $4\times$ smaller model size and comparable real-device latency.

\textbf{Visualization Results.} As shown in Fig.~\ref{fig:visual_sr}, AGD compression dramatically reduces the model size by over 40 times, without sacrificing the visual quality. The real device latency is also shrunk by 10-18 times on the four sets.

\begin{table}[htb]
  \centering
  \vspace{-1em}
  \caption{Searched architecture (operator, width) by AGD on visualization-oriented ESRGAN, where we search for five operators and their widths in the five Residual-in-Residual blocks. }
  \vspace{0.5em}
  \resizebox{0.48\textwidth}{!}{
    \begin{tabular}{c|ccccc}
    \hline
    RiR Block ID & OP1 & OP2 & OP3 & OP4 & OP5 \\ \hline
    1 & (Conv1x1, 24) & (Conv1x1, 24) & (Conv1x1, 24) & (Conv1x1, 24) & (Conv1x1, 64) \\ \hline
    2 & (DwsBlock, 24) & (DwsBlock, 24) & (ResBlock, 64) & (Conv3x3, 24) & (ResBlock, 64) \\ \hline
    3 & (Conv1x1, 24) & (Conv1x1, 24) & (Conv1x1, 24) & (Conv1x1, 24) & (Conv1x1, 64) \\ \hline
    4 & (Conv3x3, 32) & (DwsBlock, 32) & (DwsBlock, 24) & (DwsBlock, 24) & (ResBlock, 64) \\ \hline
    5 & (Conv1x1, 24) & (Conv1x1, 24) & (Conv1x1, 24) & (Conv1x1, 24) & (Conv1x1, 64) \\ \hline
    \end{tabular}
    }
  \label{table:esrgan_arch}%
  \vspace{-0.5em}
\end{table}%

\textbf{AGD Searched Architecture.} As shown in Table~\ref{table:esrgan_arch}, we find that among the original five RiR blocks, AGD assigns more computationally powerful operators to the 2nd and 4th one, while keeping the other three simple by only using Conv$1\times 1$ operations, achieving a proper balance between heavy and efficient operations.

\begin{table}[htbp]
  \centering
  \vspace{-1.5em}
  \caption{Searched architecture (operator, width) by our proposed AGD on the PSNR-oriented ESRGAN~\cite{wang2018esrgan}, where RiR denotes a Residual-in-Residual block using the operators in our search space. We search for five operators and their widths in each RiR block (totally five RiR blocks). }
  \vspace{0.5em}
  \resizebox{0.48\textwidth}{!}{
    \begin{tabular}{c|ccccc}
        \hline
        \textbf{RiR Block ID} & \textbf{OP1} & \textbf{OP2} & \textbf{OP3} & \textbf{OP4} & \textbf{OP5} \\ \hline
        1 & (DwsBlock, 32) & (DwsBlock, 32) & (DwsBlock, 48) & (DwsBlock, 24) & (ResBlock, 64) \\ \hline
        2 & (DwsBlock, 24) & (ResBlock, 32) & (Conv1x1, 32) & (Conv1x1, 24) & (Conv3x3, 64) \\ \hline
        3 & (Conv1x1, 24) & (Conv1x1, 24) & (Conv1x1, 24) & (Conv1x1, 24) & (Conv1x1, 64) \\ \hline
        4 & (DwsBlock, 32) & (DwsBlock, 48) & (DwsBlock, 24) & (DwsBlock, 24) & (ResBlock, 64) \\ \hline
        5 & (Conv1x1, 24) & (Conv1x1, 24) & (Conv1x1, 24) & (Conv1x1, 24) & (Conv1x1, 64) \\ \hline
    \end{tabular}
    }
  \label{table:esrgan_psnr}%
  \vspace{-0.5em}
\end{table}%

\textbf{Comparison with PSNR-oriented SR Methods.}
We also search for a new architecture as shown in Table~\ref{table:esrgan_psnr} under the distillation of a PSNR-oriented ESRGAN with the content loss $L_c$ only in Eq.~\ref{eqn:loss}  (following~\cite{wang2018esrgan}) within the same search space. 
The quantitative results of the original ESRGAN~\cite{wang2018esrgan}, the searched architecture of AGD, and the PSNR-oriented model VSDR~\cite{kim2016accurate} are shown in Table~\ref{table:exp_psnr}. Compared with VDSR, AGD achieves better PSNR (up to 0.44) on all the four datasets with 84.1\% fewer FLOPs and 32.6\% smaller model size. 

\vspace{-0.5em}
\section{Conclusion}
We propose the AGD framework to significantly push forward the frontier of GAN compression by introducing AutoML here. Starting with a specially designed search space of efficient building blocks, AGD performs differential neural architecture search under both the guidance of knowledge distillation and the constraint of computational resources. 
AGD is automatic with no assumptions about GAN structures, loss forms, or the availability of discriminators, and can be applicable to various GAN tasks. 
Experiments show that AGD outperforms existing options with aggressively reduced FLOPs, model size, and real-device latency, without degrading model performance. Our future work include evaluating AGD in more GAN-based applications such as data augmentation \cite{zhang2019dada}.

\begin{table}[htbp]
  \centering
  \vspace{-1em}
  \caption{Quantitative comparison with the state-of-the-art PSNR-oriented SR models. FLOPs are calculated as processing a 256$\times$256 low resolution image with a scale factor of four.}
  \vspace{0.5em}
  \resizebox{0.48\textwidth}{!}{
    \begin{tabular}{ccccccc}
    \toprule
    \multirow{2}[4]{*}{Model} & \multicolumn{1}{c}{\multirow{2}[4]{*}{GFLOPs (256x256)}} & \multicolumn{1}{c}{\multirow{2}[4]{*}{Memory (MB)}} & \multicolumn{4}{c}{PSNR} \\
\cmidrule{4-7}    \multicolumn{1}{c}{} &        &        & \multicolumn{1}{c}{Set5} & \multicolumn{1}{c}{Set14} & \multicolumn{1}{c}{BSD100} & \multicolumn{1}{c}{Urban100} \\
    \midrule
    ESRGAN & 1176.61 & 66.8   & 32.73  & 28.99  & 27.85  & 27.03 \\
    \midrule
    VDSR   & 699.36 & 2.67   & 31.35  & 28.01  & 27.29  & 25.18 \\
    \midrule
    AGD & 110.9  & 1.8    & 31.79  & 28.36  & 27.41  & 25.55 \\
    \bottomrule
    \end{tabular}%
    }
  \label{table:exp_psnr}%
  \vspace{-1em}
\end{table}%

\section*{Acknowledgements}
The work is supported by the National Science Foundation (NSF) through the Energy, Power, Control, and Networks  (EPCN) program (Award number: 1934755, 1934767).

\bibliography{reference}

\begin{thebibliography}{71}
\providecommand{\natexlab}[1]{#1}
\providecommand{\url}[1]{\texttt{#1}}
\expandafter\ifx\csname urlstyle\endcsname\relax
  \providecommand{\doi}[1]{doi: #1}\else
  \providecommand{\doi}{doi: \begingroup \urlstyle{rm}\Url}\fi

\bibitem[Aly \& Dubois(2005)Aly and Dubois]{aly2005image}
Aly, H.~A. and Dubois, E.
\newblock Image up-sampling using total-variation regularization with a new
  observation model.
\newblock \emph{IEEE Transactions on Image Processing}, 14\penalty0
  (10):\penalty0 1647--1659, 2005.

\bibitem[Banner et~al.(2018)Banner, Hubara, Hoffer, and
  Soudry]{banner2018scalable}
Banner, R., Hubara, I., Hoffer, E., and Soudry, D.
\newblock Scalable methods for 8-bit training of neural networks.
\newblock In \emph{Advances in neural information processing systems}, pp.\
  5145--5153, 2018.

\bibitem[Bevilacqua et~al.(2012)Bevilacqua, Roumy, Guillemot, and
  Alberi-Morel]{bevilacqua2012low}
Bevilacqua, M., Roumy, A., Guillemot, C., and Alberi-Morel, M.~L.
\newblock Low-complexity single-image super-resolution based on nonnegative
  neighbor embedding.
\newblock 2012.

\bibitem[Brock et~al.(2018)Brock, Donahue, and Simonyan]{brock2018large}
Brock, A., Donahue, J., and Simonyan, K.
\newblock Large scale gan training for high fidelity natural image synthesis.
\newblock \emph{arXiv preprint arXiv:1809.11096}, 2018.

\bibitem[Bul{\`o} et~al.(2016)Bul{\`o}, Porzi, and
  Kontschieder]{bulo2016dropout}
Bul{\`o}, S.~R., Porzi, L., and Kontschieder, P.
\newblock Dropout distillation.
\newblock In \emph{International Conference on Machine Learning}, pp.\
  99--107, 2016.

\bibitem[Cai et~al.(2018)Cai, Zhu, and Han]{cai2018proxylessnas}
Cai, H., Zhu, L., and Han, S.
\newblock Proxylessnas: Direct neural architecture search on target task and
  hardware.
\newblock \emph{arXiv preprint arXiv:1812.00332}, 2018.

\bibitem[Chen et~al.(2019{\natexlab{a}})Chen, Wang, Xu, Yang, Liu, Shi, Xu, Xu,
  and Tian]{chen2019data}
Chen, H., Wang, Y., Xu, C., Yang, Z., Liu, C., Shi, B., Xu, C., Xu, C., and
  Tian, Q.
\newblock Data-free learning of student networks.
\newblock In \emph{Proceedings of the IEEE International Conference on Computer
  Vision}, pp.\  3514--3522, 2019{\natexlab{a}}.

\bibitem[Chen et~al.(2020{\natexlab{a}})Chen, Wang, Shu, Tang, Xu, Shi, Xu,
  Tian, and Xu]{chen2020frequency}
Chen, H., Wang, Y., Shu, H., Tang, Y., Xu, C., Shi, B., Xu, C., Tian, Q., and
  Xu, C.
\newblock Frequency domain compact 3d convolutional neural networks.
\newblock In \emph{Proceedings of the IEEE/CVF Conference on Computer Vision
  and Pattern Recognition}, pp.\  1641--1650, 2020{\natexlab{a}}.

\bibitem[Chen et~al.(2020{\natexlab{b}})Chen, Wang, Shu, Wen, Xu, Shi, Xu, and
  Xu]{chen2020distilling}
Chen, H., Wang, Y., Shu, H., Wen, C., Xu, C., Shi, B., Xu, C., and Xu, C.
\newblock Distilling portable generative adversarial networks for image
  translation.
\newblock \emph{arXiv preprint arXiv:2003.03519}, 2020{\natexlab{b}}.

\bibitem[Chen et~al.(2018{\natexlab{a}})Chen, Collins, Zhu, Papandreou, Zoph,
  Schroff, Adam, and Shlens]{chen2018searching}
Chen, L.-C., Collins, M., Zhu, Y., Papandreou, G., Zoph, B., Schroff, F., Adam,
  H., and Shlens, J.
\newblock Searching for efficient multi-scale architectures for dense image
  prediction.
\newblock In \emph{Advances in neural information processing systems}, pp.\
  8699--8710, 2018{\natexlab{a}}.

\bibitem[Chen et~al.(2019{\natexlab{b}})Chen, Gong, Liu, Zhang, Li, and
  Wang]{chen2019fasterseg}
Chen, W., Gong, X., Liu, X., Zhang, Q., Li, Y., and Wang, Z.
\newblock Fasterseg: Searching for faster real-time semantic segmentation.
\newblock \emph{arXiv preprint arXiv:1912.10917}, 2019{\natexlab{b}}.

\bibitem[Chen et~al.(2018{\natexlab{b}})Chen, Lai, and Liu]{chen2018cartoongan}
Chen, Y., Lai, Y.-K., and Liu, Y.-J.
\newblock Cartoongan: Generative adversarial networks for photo cartoonization.
\newblock In \emph{Proceedings of the IEEE conference on computer vision and
  pattern recognition}, pp.\  9465--9474, 2018{\natexlab{b}}.

\bibitem[Cheng et~al.(2018)Cheng, Lin, Juan, Wei, and Sun]{cheng2018instanas}
Cheng, A.-C., Lin, C.~H., Juan, D.-C., Wei, W., and Sun, M.
\newblock Instanas: Instance-aware neural architecture search.
\newblock \emph{arXiv preprint arXiv:1811.10201}, 2018.

\bibitem[Courbariaux et~al.(2015)Courbariaux, Bengio, and
  David]{courbariaux2015binaryconnect}
Courbariaux, M., Bengio, Y., and David, J.-P.
\newblock {BinaryConnect}: Training deep neural networks with binary weights
  during propagations.
\newblock In \emph{Advances in Neural Information Processing Systems}, pp.\
  3123--3131, 2015.

\bibitem[Elsken et~al.(2018)Elsken, Metzen, and Hutter]{elsken2018neural}
Elsken, T., Metzen, J.~H., and Hutter, F.
\newblock Neural architecture search: A survey.
\newblock \emph{arXiv preprint arXiv:1808.05377}, 2018.

\bibitem[Gong et~al.(2019)Gong, Chang, Jiang, and Wang]{gong2019autogan}
Gong, X., Chang, S., Jiang, Y., and Wang, Z.
\newblock Autogan: Neural architecture search for generative adversarial
  networks.
\newblock In \emph{Proceedings of the IEEE International Conference on Computer
  Vision}, pp.\  3224--3234, 2019.

\bibitem[Gong et~al.(2014)Gong, Liu, Yang, and Bourdev]{gong2014compressing}
Gong, Y., Liu, L., Yang, M., and Bourdev, L.
\newblock Compressing deep convolutional networks using vector quantization.
\newblock \emph{arXiv preprint arXiv:1412.6115}, 2014.

\bibitem[Goodfellow et~al.(2014)Goodfellow, Pouget-Abadie, Mirza, Xu,
  Warde-Farley, Ozair, Courville, and Bengio]{goodfellow2014generative}
Goodfellow, I., Pouget-Abadie, J., Mirza, M., Xu, B., Warde-Farley, D., Ozair,
  S., Courville, A., and Bengio, Y.
\newblock Generative adversarial nets.
\newblock In \emph{Advances in neural information processing systems}, pp.\
  2672--2680, 2014.

\bibitem[Gui et~al.(2016)Gui, Sun, Ji, Tao, and Tan]{gui2016feature}
Gui, J., Sun, Z., Ji, S., Tao, D., and Tan, T.
\newblock Feature selection based on structured sparsity: A comprehensive
  study.
\newblock \emph{IEEE transactions on neural networks and learning systems},
  28\penalty0 (7):\penalty0 1490--1507, 2016.

\bibitem[Gui et~al.(2020)Gui, Sun, Wen, Tao, and Ye]{gui2020review}
Gui, J., Sun, Z., Wen, Y., Tao, D., and Ye, J.
\newblock A review on generative adversarial networks: Algorithms, theory, and
  applications.
\newblock \emph{arXiv preprint arXiv:2001.06937}, 2020.

\bibitem[Gulrajani et~al.(2017)Gulrajani, Ahmed, Arjovsky, Dumoulin, and
  Courville]{gulrajani2017improved}
Gulrajani, I., Ahmed, F., Arjovsky, M., Dumoulin, V., and Courville, A.~C.
\newblock Improved training of wasserstein gans.
\newblock In \emph{Advances in neural information processing systems}, pp.\
  5767--5777, 2017.

\bibitem[Gumbel(1948)]{gumbel1948statistical}
Gumbel, E.~J.
\newblock \emph{Statistical theory of extreme values and some practical
  applications: a series of lectures}, volume~33.
\newblock US Government Printing Office, 1948.

\bibitem[Han et~al.(2020)Han, Wang, Tian, Guo, Xu, and Xu]{han2020ghostnet}
Han, K., Wang, Y., Tian, Q., Guo, J., Xu, C., and Xu, C.
\newblock Ghostnet: More features from cheap operations.
\newblock In \emph{Proceedings of the IEEE/CVF Conference on Computer Vision
  and Pattern Recognition}, pp.\  1580--1589, 2020.

\bibitem[Han et~al.(2015)Han, Mao, and Dally]{han2015deep}
Han, S., Mao, H., and Dally, W.~J.
\newblock Deep compression: Compressing deep neural networks with pruning,
  trained quantization and huffman coding.
\newblock \emph{arXiv preprint arXiv:1510.00149}, 2015.

\bibitem[He et~al.(2017)He, Zhang, and Sun]{he2017channel}
He, Y., Zhang, X., and Sun, J.
\newblock Channel pruning for accelerating very deep neural networks.
\newblock In \emph{Proceedings of the IEEE International Conference on Computer
  Vision}, pp.\  1389--1397, 2017.

\bibitem[He et~al.(2018)He, Lin, Liu, Wang, Li, and Han]{he2018amc}
He, Y., Lin, J., Liu, Z., Wang, H., Li, L.-J., and Han, S.
\newblock Amc: Automl for model compression and acceleration on mobile devices.
\newblock In \emph{Proceedings of the European Conference on Computer Vision
  (ECCV)}, pp.\  784--800, 2018.

\bibitem[Heusel et~al.(2017)Heusel, Ramsauer, Unterthiner, Nessler, and
  Hochreiter]{heusel2017gans}
Heusel, M., Ramsauer, H., Unterthiner, T., Nessler, B., and Hochreiter, S.
\newblock Gans trained by a two time-scale update rule converge to a local nash
  equilibrium.
\newblock In \emph{Advances in Neural Information Processing Systems}, pp.\
  6626--6637, 2017.

\bibitem[Hinton et~al.(2015)Hinton, Vinyals, and Dean]{hinton2015distilling}
Hinton, G., Vinyals, O., and Dean, J.
\newblock Distilling the knowledge in a neural network.
\newblock \emph{arXiv preprint arXiv:1503.02531}, 2015.

\bibitem[Howard et~al.(2019)Howard, Sandler, Chu, Chen, Chen, Tan, Wang, Zhu,
  Pang, Vasudevan, et~al.]{howard2019searching}
Howard, A., Sandler, M., Chu, G., Chen, L.-C., Chen, B., Tan, M., Wang, W.,
  Zhu, Y., Pang, R., Vasudevan, V., et~al.
\newblock Searching for mobilenetv3.
\newblock In \emph{Proceedings of the IEEE International Conference on Computer
  Vision}, pp.\  1314--1324, 2019.

\bibitem[Huang et~al.(2015)Huang, Singh, and Ahuja]{huang2015single}
Huang, J.-B., Singh, A., and Ahuja, N.
\newblock Single image super-resolution from transformed self-exemplars.
\newblock In \emph{Proceedings of the IEEE Conference on Computer Vision and
  Pattern Recognition}, pp.\  5197--5206, 2015.

\bibitem[Hutter et~al.(2019)Hutter, Kotthoff, and
  Vanschoren]{hutter2019automated}
Hutter, F., Kotthoff, L., and Vanschoren, J.
\newblock \emph{Automated Machine Learning}.
\newblock Springer, 2019.

\bibitem[Jacob et~al.(2018)Jacob, Kligys, Chen, Zhu, Tang, Howard, Adam, and
  Kalenichenko]{jacob2018quantization}
Jacob, B., Kligys, S., Chen, B., Zhu, M., Tang, M., Howard, A., Adam, H., and
  Kalenichenko, D.
\newblock Quantization and training of neural networks for efficient
  integer-arithmetic-only inference.
\newblock In \emph{Proceedings of the IEEE Conference on Computer Vision and
  Pattern Recognition}, pp.\  2704--2713, 2018.

\bibitem[Jiang et~al.(2019)Jiang, Gong, Liu, Cheng, Fang, Shen, Yang, Zhou, and
  Wang]{jiang2019enlightengan}
Jiang, Y., Gong, X., Liu, D., Cheng, Y., Fang, C., Shen, X., Yang, J., Zhou,
  P., and Wang, Z.
\newblock Enlightengan: Deep light enhancement without paired supervision.
\newblock \emph{arXiv preprint arXiv:1906.06972}, 2019.

\bibitem[Johnson et~al.(2016)Johnson, Alahi, and
  Fei-Fei]{johnson2016perceptual}
Johnson, J., Alahi, A., and Fei-Fei, L.
\newblock Perceptual losses for real-time style transfer and super-resolution.
\newblock In \emph{European conference on computer vision}, pp.\  694--711.
  Springer, 2016.

\bibitem[Karras et~al.(2017)Karras, Aila, Laine, and
  Lehtinen]{karras2017progressive}
Karras, T., Aila, T., Laine, S., and Lehtinen, J.
\newblock Progressive growing of gans for improved quality, stability, and
  variation.
\newblock \emph{arXiv preprint arXiv:1710.10196}, 2017.

\bibitem[Kim et~al.(2016)Kim, Kwon~Lee, and Mu~Lee]{kim2016accurate}
Kim, J., Kwon~Lee, J., and Mu~Lee, K.
\newblock Accurate image super-resolution using very deep convolutional
  networks.
\newblock In \emph{Proceedings of the IEEE conference on computer vision and
  pattern recognition}, pp.\  1646--1654, 2016.

\bibitem[Kupyn et~al.(2019)Kupyn, Martyniuk, Wu, and Wang]{kupyn2019deblurgan}
Kupyn, O., Martyniuk, T., Wu, J., and Wang, Z.
\newblock Deblurgan-v2: Deblurring (orders-of-magnitude) faster and better.
\newblock In \emph{Proceedings of the IEEE International Conference on Computer
  Vision}, pp.\  8878--8887, 2019.

\bibitem[Ledig et~al.(2017)Ledig, Theis, Husz{\'a}r, Caballero, Cunningham,
  Acosta, Aitken, Tejani, Totz, Wang, et~al.]{ledig2017photo}
Ledig, C., Theis, L., Husz{\'a}r, F., Caballero, J., Cunningham, A., Acosta,
  A., Aitken, A., Tejani, A., Totz, J., Wang, Z., et~al.
\newblock Photo-realistic single image super-resolution using a generative
  adversarial network.
\newblock In \emph{Proceedings of the IEEE conference on computer vision and
  pattern recognition}, pp.\  4681--4690, 2017.

\bibitem[Li et~al.(2016)Li, Kadav, Durdanovic, Samet, and Graf]{li2016pruning}
Li, H., Kadav, A., Durdanovic, I., Samet, H., and Graf, H.~P.
\newblock Pruning filters for efficient convnets.
\newblock \emph{arXiv preprint arXiv:1608.08710}, 2016.

\bibitem[Li et~al.(2020)Li, Lin, Ding, Liu, Zhu, and Han]{li2020gan}
Li, M., Lin, J., Ding, Y., Liu, Z., Zhu, J.-Y., and Han, S.
\newblock Gan compression: Efficient architectures for interactive conditional
  gans.
\newblock In \emph{Proceedings of the IEEE/CVF Conference on Computer Vision
  and Pattern Recognition}, pp.\  5284--5294, 2020.

\bibitem[Liu et~al.(2019)Liu, Chen, Schroff, Adam, Hua, Yuille, and
  Fei-Fei]{liu2019auto}
Liu, C., Chen, L.-C., Schroff, F., Adam, H., Hua, W., Yuille, A.~L., and
  Fei-Fei, L.
\newblock Auto-deeplab: Hierarchical neural architecture search for semantic
  image segmentation.
\newblock In \emph{Proceedings of the IEEE Conference on Computer Vision and
  Pattern Recognition}, pp.\  82--92, 2019.

\bibitem[Liu et~al.(2018{\natexlab{a}})Liu, Simonyan, and Yang]{liu2018darts}
Liu, H., Simonyan, K., and Yang, Y.
\newblock Darts: Differentiable architecture search.
\newblock \emph{arXiv preprint arXiv:1806.09055}, 2018{\natexlab{a}}.

\bibitem[Liu et~al.(2018{\natexlab{b}})Liu, Lin, Zhou, Nan, Liu, and
  Du]{liu2018adadeep}
Liu, S., Lin, Y., Zhou, Z., Nan, K., Liu, H., and Du, J.
\newblock {On-demand deep model compression for mobile devices: A usage-driven
  model selection framework}.
\newblock In \emph{Proceedings of International Conference on Mobile Systems,
  Applications, and Services}, pp.\  389--400. ACM, 2018{\natexlab{b}}.

\bibitem[Lopez-Paz et~al.(2015)Lopez-Paz, Bottou, Sch{\"o}lkopf, and
  Vapnik]{lopez2015unifying}
Lopez-Paz, D., Bottou, L., Sch{\"o}lkopf, B., and Vapnik, V.
\newblock Unifying distillation and privileged information.
\newblock \emph{arXiv preprint arXiv:1511.03643}, 2015.

\bibitem[Luo et~al.(2017)Luo, Wu, and Lin]{luo2017thinet}
Luo, J.-H., Wu, J., and Lin, W.
\newblock Thinet: A filter level pruning method for deep neural network
  compression.
\newblock In \emph{Proceedings of the IEEE international conference on computer
  vision}, pp.\  5058--5066, 2017.

\bibitem[Ma et~al.(2018)Ma, Zhang, Zheng, and Sun]{ma2018shufflenet}
Ma, N., Zhang, X., Zheng, H.-T., and Sun, J.
\newblock Shufflenet v2: Practical guidelines for efficient cnn architecture
  design.
\newblock In \emph{Proceedings of the European Conference on Computer Vision
  (ECCV)}, pp.\  116--131, 2018.

\bibitem[Martin et~al.(2001)Martin, Fowlkes, Tal, Malik,
  et~al.]{martin2001database}
Martin, D., Fowlkes, C., Tal, D., Malik, J., et~al.
\newblock A database of human segmented natural images and its application to
  evaluating segmentation algorithms and measuring ecological statistics.
\newblock Iccv Vancouver:, 2001.

\bibitem[Miyato et~al.(2018)Miyato, Kataoka, Koyama, and
  Yoshida]{miyato2018spectral}
Miyato, T., Kataoka, T., Koyama, M., and Yoshida, Y.
\newblock Spectral normalization for generative adversarial networks.
\newblock \emph{arXiv preprint arXiv:1802.05957}, 2018.

\bibitem[{NVIDIA Inc.}()]{2080ti}
{NVIDIA Inc.}
\newblock {NVIDIA GEFORCE RTX 2080 Ti}.
\newblock
  \url{https://www.nvidia.com/en-us/geforce/graphics-cards/rtx-2080-ti/},
  accessed 2020-06-14.

\bibitem[Polino et~al.(2018)Polino, Pascanu, and Alistarh]{polino2018model}
Polino, A., Pascanu, R., and Alistarh, D.
\newblock Model compression via distillation and quantization.
\newblock \emph{arXiv preprint arXiv:1802.05668}, 2018.

\bibitem[Rastegari et~al.(2016)Rastegari, Ordonez, Redmon, and
  Farhadi]{rastegari2016xnor}
Rastegari, M., Ordonez, V., Redmon, J., and Farhadi, A.
\newblock {XNOR-Net}: {ImageNet} classification using binary convolutional
  neural networks.
\newblock In \emph{European Conference on Computer Vision}, pp.\  525--542,
  2016.

\bibitem[Sanakoyeu et~al.(2018)Sanakoyeu, Kotovenko, Lang, and
  Ommer]{sanakoyeu2018style}
Sanakoyeu, A., Kotovenko, D., Lang, S., and Ommer, B.
\newblock A style-aware content loss for real-time hd style transfer.
\newblock In \emph{Proceedings of the European Conference on Computer Vision
  (ECCV)}, pp.\  698--714, 2018.

\bibitem[Shi et~al.(2016)Shi, Caballero, Husz{\'a}r, Totz, Aitken, Bishop,
  Rueckert, and Wang]{shi2016real}
Shi, W., Caballero, J., Husz{\'a}r, F., Totz, J., Aitken, A.~P., Bishop, R.,
  Rueckert, D., and Wang, Z.
\newblock Real-time single image and video super-resolution using an efficient
  sub-pixel convolutional neural network.
\newblock In \emph{Proceedings of the IEEE conference on computer vision and
  pattern recognition}, pp.\  1874--1883, 2016.

\bibitem[Shu et~al.(2019)Shu, Wang, Jia, Han, Chen, Xu, Tian, and
  Xu]{shu2019co}
Shu, H., Wang, Y., Jia, X., Han, K., Chen, H., Xu, C., Tian, Q., and Xu, C.
\newblock Co-evolutionary compression for unpaired image translation.
\newblock In \emph{Proceedings of the IEEE International Conference on Computer
  Vision}, pp.\  3235--3244, 2019.

\bibitem[Tan \& Le(2019)Tan and Le]{tan2019efficientnet}
Tan, M. and Le, Q.~V.
\newblock Efficientnet: Rethinking model scaling for convolutional neural
  networks.
\newblock \emph{arXiv preprint arXiv:1905.11946}, 2019.

\bibitem[Tan et~al.(2019)Tan, Chen, Pang, Vasudevan, Sandler, Howard, and
  Le]{tan2019mnasnet}
Tan, M., Chen, B., Pang, R., Vasudevan, V., Sandler, M., Howard, A., and Le,
  Q.~V.
\newblock Mnasnet: Platform-aware neural architecture search for mobile.
\newblock In \emph{Proceedings of the IEEE Conference on Computer Vision and
  Pattern Recognition}, pp.\  2820--2828, 2019.

\bibitem[Timofte et~al.(2017)Timofte, Agustsson, Van~Gool, Yang, and
  Zhang]{Agustsson_2017_CVPR_Workshops}
Timofte, R., Agustsson, E., Van~Gool, L., Yang, M.-H., and Zhang, L.
\newblock Ntire 2017 challenge on single image super-resolution: Methods and
  results.
\newblock In \emph{Proceedings of the IEEE conference on computer vision and
  pattern recognition workshops}, pp.\  114--125, 2017.

\bibitem[Ulyanov et~al.(2016)Ulyanov, Vedaldi, and
  Lempitsky]{ulyanov2016instance}
Ulyanov, D., Vedaldi, A., and Lempitsky, V.
\newblock Instance normalization: The missing ingredient for fast stylization.
\newblock \emph{arXiv preprint arXiv:1607.08022}, 2016.

\bibitem[Wang et~al.(2018{\natexlab{a}})Wang, Yu, Wu, Gu, Liu, Dong, Qiao, and
  Change~Loy]{wang2018esrgan}
Wang, X., Yu, K., Wu, S., Gu, J., Liu, Y., Dong, C., Qiao, Y., and Change~Loy,
  C.
\newblock Esrgan: Enhanced super-resolution generative adversarial networks.
\newblock In \emph{Proceedings of the European Conference on Computer Vision
  (ECCV)}, pp.\  0--0, 2018{\natexlab{a}}.

\bibitem[Wang et~al.(2018{\natexlab{b}})Wang, Nguyen, Zhao, Wang, Lin, and
  Baraniuk]{energynet}
Wang, Y., Nguyen, T., Zhao, Y., Wang, Z., Lin, Y., and Baraniuk, R.
\newblock Energynet: Energy-efficient dynamic inference.
\newblock \emph{NeurIPS workshop}, 2018{\natexlab{b}}.

\bibitem[Wang et~al.(2018{\natexlab{c}})Wang, Xu, Xu, and
  Tao]{wang2018adversarial}
Wang, Y., Xu, C., Xu, C., and Tao, D.
\newblock Adversarial learning of portable student networks.
\newblock In \emph{Thirty-Second AAAI Conference on Artificial Intelligence},
  2018{\natexlab{c}}.

\bibitem[Wen et~al.(2016)Wen, Wu, Wang, Chen, and Li]{wen2016learning}
Wen, W., Wu, C., Wang, Y., Chen, Y., and Li, H.
\newblock Learning structured sparsity in deep neural networks.
\newblock In \emph{Advances in neural information processing systems}, pp.\
  2074--2082, 2016.

\bibitem[Wu et~al.(2019)Wu, Dai, Zhang, Wang, Sun, Wu, Tian, Vajda, Jia, and
  Keutzer]{wu2019fbnet}
Wu, B., Dai, X., Zhang, P., Wang, Y., Sun, F., Wu, Y., Tian, Y., Vajda, P.,
  Jia, Y., and Keutzer, K.
\newblock Fbnet: Hardware-aware efficient convnet design via differentiable
  neural architecture search.
\newblock In \emph{Proceedings of the IEEE Conference on Computer Vision and
  Pattern Recognition}, pp.\  10734--10742, 2019.

\bibitem[Wu et~al.(2016)Wu, Leng, Wang, Hu, and Cheng]{wu2016quantized}
Wu, J., Leng, C., Wang, Y., Hu, Q., and Cheng, J.
\newblock Quantized convolutional neural networks for mobile devices.
\newblock In \emph{IEEE Conference on Computer Vision and Pattern Recognition},
  pp.\  4820--4828, 2016.

\bibitem[Wu et~al.(2018)Wu, Wang, Wu, Wang, Veeraraghavan, and Lin]{wu2018deep}
Wu, J., Wang, Y., Wu, Z., Wang, Z., Veeraraghavan, A., and Lin, Y.
\newblock Deep k-means: Re-training and parameter sharing with harder cluster
  assignments for compressing deep convolutions.
\newblock In \emph{International Conference on Machine Learning}, pp.\
  5359--5368, 2018.

\bibitem[Yang et~al.(2019)Yang, Wang, Wang, Xu, Liu, and
  Guo]{yang2019controllable}
Yang, S., Wang, Z., Wang, Z., Xu, N., Liu, J., and Guo, Z.
\newblock Controllable artistic text style transfer via shape-matching gan.
\newblock In \emph{Proceedings of the IEEE International Conference on Computer
  Vision}, pp.\  4442--4451, 2019.

\bibitem[Yu \& Huang(2019)Yu and Huang]{yu2019universally}
Yu, J. and Huang, T.~S.
\newblock Universally slimmable networks and improved training techniques.
\newblock In \emph{Proceedings of the IEEE International Conference on Computer
  Vision}, pp.\  1803--1811, 2019.

\bibitem[Zeyde et~al.(2010)Zeyde, Elad, and Protter]{zeyde2010single}
Zeyde, R., Elad, M., and Protter, M.
\newblock On single image scale-up using sparse-representations.
\newblock In \emph{International conference on curves and surfaces}, pp.\
  711--730. Springer, 2010.

\bibitem[Zhang et~al.(2019)Zhang, Wang, Liu, and Ling]{zhang2019dada}
Zhang, X., Wang, Z., Liu, D., and Ling, Q.
\newblock Dada: Deep adversarial data augmentation for extremely low data
  regime classification.
\newblock In \emph{ICASSP 2019-2019 IEEE International Conference on Acoustics,
  Speech and Signal Processing (ICASSP)}, pp.\  2807--2811. IEEE, 2019.

\bibitem[Zhu et~al.(2017)Zhu, Park, Isola, and Efros]{zhu2017unpaired}
Zhu, J.-Y., Park, T., Isola, P., and Efros, A.~A.
\newblock Unpaired image-to-image translation using cycle-consistent
  adversarial networks.
\newblock In \emph{Proceedings of the IEEE international conference on computer
  vision}, pp.\  2223--2232, 2017.

\bibitem[Zoph \& Le(2016)Zoph and Le]{zoph2016neural}
Zoph, B. and Le, Q.~V.
\newblock Neural architecture search with reinforcement learning.
\newblock \emph{arXiv preprint arXiv:1611.01578}, 2016.

\end{thebibliography}
\bibliographystyle{icml2020}

\end{document}